%% file: Main_EN_TITS_ver3.tex
\documentclass[letterpaper,journal]{IEEEtran}
\usepackage[T1]{fontenc}
\usepackage{amsmath,amssymb,amsfonts,upgreek}
\allowdisplaybreaks[1]
\usepackage{graphicx,booktabs,array,tabularx,multirow,makecell,threeparttable}
\usepackage{xcolor}
\newif\ifshowrevisions
\showrevisionsfalse
\newcommand{\rev}[1]{{\ifshowrevisions\color{red}\fi#1}}
\usepackage[caption=false,font=footnotesize]{subfig}
\usepackage{cite}
\usepackage[hyphens]{url}
\usepackage{stfloats}
\usepackage[hidelinks]{hyperref}
\usepackage{orcidlink}
\newcolumntype{Y}{>{\raggedright\arraybackslash}X}
\begin{document}
\raggedbottom

\title{A Sequential Framework for Urban Drone Delivery: Network Design, Vehicle Sizing, and Energy and Cost Assessment}

%\title{Linking Network Design and Drone Sizing to Energy Consumption and Operating Cost in Urban Delivery}

\author{Hong-Su~Nam~\orcidlink{0000-0003-2360-3990},
Seokhwan~Lee~\orcidlink{0009-0005-2983-7722},
Jungwoo~Cho~\orcidlink{0000-0003-0770-9023},
and~Hak-Tae~Lee~\orcidlink{0000-0002-0674-9931}%
\thanks{H.-S. Nam, J. Cho, and H.-T. Lee are with the Department of Aerospace Engineering and the Program in Aerospace Systems Convergence, Inha University, Incheon 21999, Republic of Korea (e-mail: namhs501@inha.edu; jungwoo.cho@inha.ac.kr; haktae.lee@inha.ac.kr).}%
\thanks{S. Lee was with the Department of Aerospace Engineering, Inha University, Incheon 21999, Republic of Korea. He is now with the Aerospace R\&D Center, Korean Air, Daejeon 34054, Republic of Korea (e-mail: seokhwanlee@koreanair.com).}%
\thanks{J. Cho and H.-T. Lee are also with the Aerospace Systems Research Institute, Inha University, Incheon 21999, Republic of Korea. Corresponding author: Hak-Tae Lee.}}
\markboth{IEEE Transactions on Intelligent Transportation Systems}{Nam et al.: A Sequential Framework for Urban Drone Delivery}
\maketitle

% LHT - Pretty much the final abstract 2026.11.03.
\begin{abstract}
A sequential engineering framework is presented for urban drone delivery, linking hub location, route planning, mission-specific vehicle sizing, and energy and operating cost assessment. Aggregate origin–destination demand is assumed known a priori. For each hub configuration, the maximum required link distance is minimized, and feasible routes that return to their departure hubs are generated. Route frequencies are determined using mixed-integer linear programming to satisfy delivery demand. The resulting range requirement guides motor–propeller selection and battery sizing to minimize vehicle takeoff mass. Flight distances and frequencies, cruise performance, and vehicle dimensions are combined to estimate monthly energy consumption, fleet size, hub area, staffing, and operating cost. The framework is applied to district-level Seoul delivery data. Increasing the number of hubs from one to four reduces the required range from 30.6 to 21.9 km and vehicle takeoff mass from 8.12 to 6.12 kg. Despite a 21.5\% increase in total flight distance, monthly flight energy consumption decreases by 13.1\%, from 15.2 to 13.2 MWh. Battery replacement is the largest vehicle-related operating expense, while operating cost per delivery increases from USD 1.18 to USD 2.20 because of higher rent and labor costs at the hubs. Among the designs evaluated, range-matched vehicles achieve the lowest operating cost in the two- and four-hub networks; a longer-range vehicle is least costly in the three-hub network. Sensitivity analyses further examine the effects of delivery demand and operating assumptions. The results demonstrate that reductions in vehicle mass and energy consumption do not necessarily lead to lower system operating cost.
\end{abstract}

\section{Introduction}\label{sec:intro}
E-commerce and on-demand delivery have increased interest in fast and flexible last-mile transportation. Drones can operate independently of the road network and have been studied in terms of efficiency, accessibility, and sustainability\cite{moshref2021,jazairy2025,garg2023}. However, the feasibility and cost of urban drone delivery depend on numerous technical and operational factors. These include vehicle range and energy consumption, hub locations, fleet size, and facility and staffing costs. Furthermore, these factors are not independent, and changes in one may affect several others.

Quadrotor drones are considered in this study because their vertical takeoff and landing capabilities and ability to hover are advantageous for urban delivery. However, their relatively low cruise efficiency compared with fixed-wing aircraft limits their flight range. Consequently, the distances between delivery hubs and service locations become important factors in vehicle design, as longer required ranges generally lead to larger batteries and heavier vehicles.

% LHT Energy added here
For example, increasing the number of hubs may reduce the required flight range and allow lighter vehicles, but it may also increase facility and staffing costs while changing fleet requirements. Such competing effects motivate an integrated assessment of network design, vehicle sizing, energy consumption, and operating costs.

% Previous work

Drone routing research addresses cooperation between transportation modes and vehicle energy constraints. Murray and Chu\cite{murray2015} formulated a drone-assisted delivery routing problem, and Agatz et al.\cite{agatz2018} investigated optimization approaches for the traveling salesman problem with a drone. Dorling et al.\cite{dorling2017} incorporated energy consumption dependent on battery and payload mass, together with drone reuse, into route planning. Other studies have addressed cost minimization in truck--drone cooperation\cite{ha2018} and extensions of operating range using public transportation networks\cite{choudhury2021,moadab2022,huang2020}. These studies demonstrate the importance of linking feasible routes and energy use when evaluating delivery plans.

Facility and network studies relate flight range and base locations to service coverage and routing. Hong et al.\cite{hong2018} considered the coverage of recharging facilities under range restrictions, whereas Pinto and Lagorio\cite{pinto2022} addressed point-to-point drone delivery networks with intermediate charging stations. Liu\cite{liu2023} evaluated a battery-constrained depot network through optimization and simulation. Studies of facility coverage and capacity\cite{chauhan2019} and delivery routing with parcel lockers\cite{zhu2024} also connect transportation infrastructure with route planning. These contributions provide a basis for evaluating networks in which facility location and vehicle range constraints interact.

Research on urban-scale operations has expanded to corridor and route-network design\cite{he2022,he2024,he2025}, risk-aware unmanned aerial vehicle traffic networks\cite{nicholson2025}, urban airspace availability assessment \cite{cho2018,cho2021} and very-low-level traffic-density estimation\cite{doole2020}. Online demand planning\cite{paul2025}, consumer delivery preferences\cite{kim2020}, third-party ground-risk assessment\cite{choi2025}, and barriers to urban air mobility adoption\cite{sadrani2025} further highlight the role of demand, safety, and institutional conditions. 

% LHT - energy added
Doole et al.\cite{doole2020} estimated drone-based food-delivery costs of 0.40--2.51 EUR per meal order across scenarios that included annualized investment and operating costs. Bao et al.\cite{bao2026} estimated 4.17 EUR per parcel in a 200-customer-node scenario using a five-year total-cost-of-ownership model that included vehicle and battery acquisition, labor, and infrastructure costs. Differences in service units, operating assumptions, and cost coverage limit direct comparisons of these estimates. Moreover, the effects of network-dependent range requirements on component level vehicle design, and their subsequent implications for energy consumption, fleet requirements, and hub operating costs, have received comparatively limited attention.

% Building on the existing literature, an integrated engineering framework is proposed to link drone delivery network design, flight route planning, mission-specific vehicle sizing, and energy and operating cost assessment. The framework follows a sequential procedure in which the outputs of each design stage are used in subsequent analyses. In particular, network-derived range requirements are applied to both route planning and vehicle sizing, while vehicle performance and flight activity are used to estimate energy consumption, fleet size, hub area, staffing, and operating costs. Vehicle sizing and performance evaluation are based on the approaches of Jeong\cite{jeong2021}, Pollet et al.\cite{pollet2021}, and Nam et al.\cite{nam2024}.

% The framework is demonstrated using actual ground-delivery data from Seoul, Republic of Korea. Hub configurations and vehicle designs are compared under consistent operating assumptions, followed by sensitivity analyses of demand and cost parameters.

\rev{This paper presents a sequential engineering framework for assessing how network design affects vehicle sizing, energy use, and operating cost in urban drone delivery. Building on the framework introduced by Nam and Lee at the AIAA AVIATION 2026 Forum~\cite{nam2026aviation}, the present study details how hub locations determine the range required for route planning and vehicle sizing. The resulting flight plan and vehicle performance are then used to estimate fleet size, hub staffing, energy consumption, and operating cost.} Vehicle sizing and performance evaluation are based on the approaches of Jeong\cite{jeong2021}, Pollet et al.\cite{pollet2021}, and Nam et al.\cite{nam2024}.

\rev{The sizing estimates are compared with commercial drone specifications. Ground-delivery data from Seoul, Republic of Korea, provide a case study to demonstrate the framework and examine whether the cost rankings of hub configurations change with demand and operating assumptions.}

\suppressfloats[t]
\begin{figure}[t]
\centering
\includegraphics[width=0.85\linewidth]{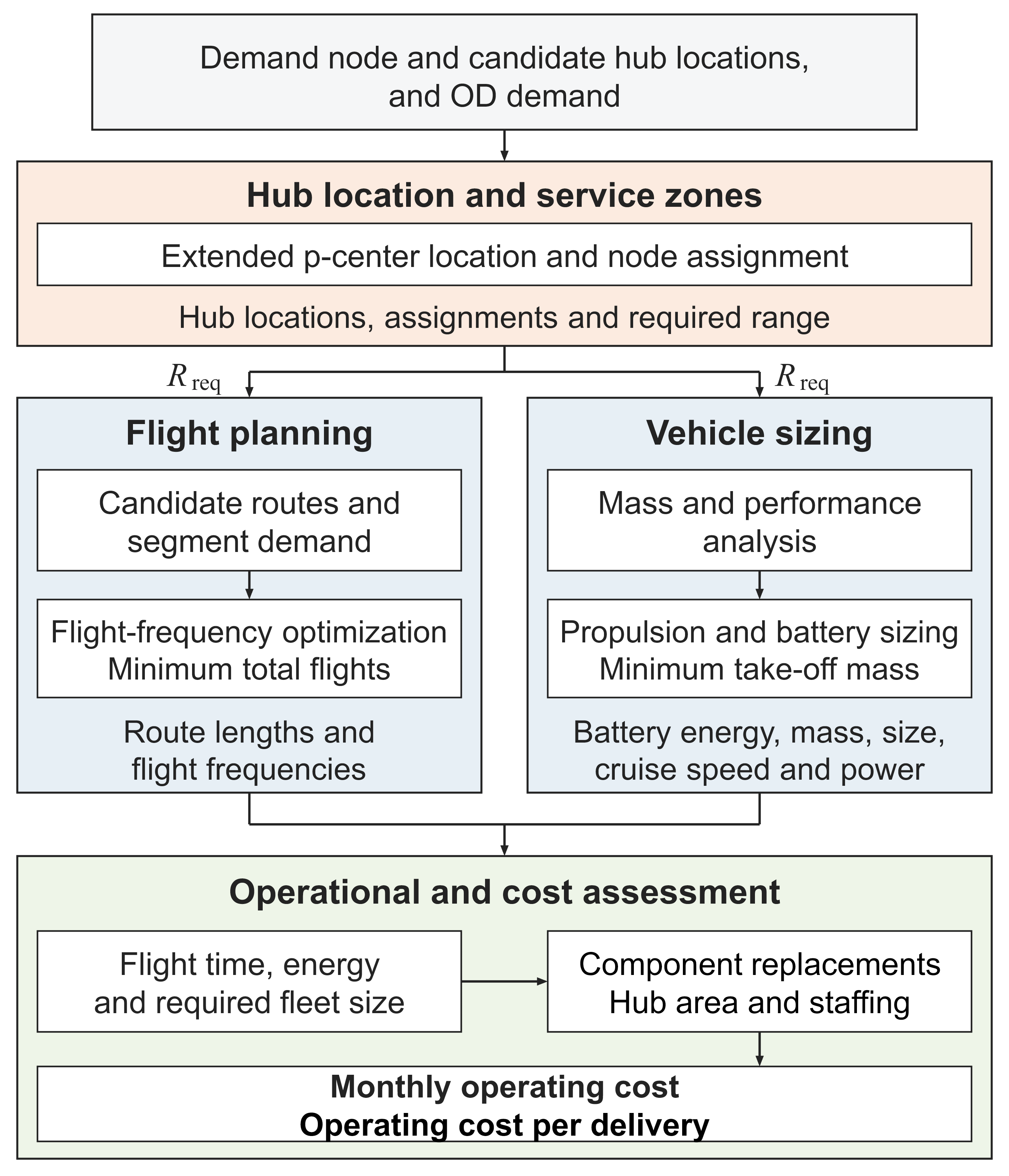}
\caption{Sequential design and evaluation framework for urban drone delivery.}
\label{fig:framework}
\end{figure}

The main contributions of this study are summarized as follows:
\begin{enumerate}
%\item \textbf{Integrated, sequential engineering framework}: 
\item A closely linked framework is established to integrate network design, flight-route optimization, component-based vehicle sizing, and energy consumption and operating cost assessment. The sequential procedure explicitly accounts for the interdependencies among these stages.

%\item \textbf{Application to an urban delivery network}: 
\item The framework is applied to Seoul using actual ground delivery data. Increasing the number of hubs from one to four reduces vehicle mass and total flight energy consumption despite an increase in total flight distance, but increases operating cost per delivery from USD 1.18 to USD 2.20. Comparisons across vehicle designs show that minimum-mass designs do not necessarily minimize operating cost because vehicle range also affects flight frequency and staffing requirements.

%\item \textbf{Sensitivity analysis}:
\item The effects of delivery demand, operating hours, component lifetimes, hub area, and staffing are examined to assess the robustness of operating cost comparisons. The one-hub configuration remains the least costly across the evaluated parameter ranges, although the relative cost rankings of multiple-hub configurations can change.
\end{enumerate}

Section~\ref{sec:framework} introduces the delivery system and the framework. Section~\ref{sec:network} presents hub location and flight-route planning, and Section~\ref{sec:vehicle} describes vehicle sizing and performance evaluation. Section~\ref{sec:cost} uses their outputs to calculate energy consumption, fleet size, hub space, staffing, and operating costs. Section~\ref{sec:results} presents the Seoul baseline, comparisons across vehicle designs, propulsion selection, and sensitivity results. Section~\ref{sec:conclusions} concludes the paper.

\section{Delivery System and Analysis Framework}\label{sec:framework}

The proposed framework connects delivery network design, flight-route planning, vehicle sizing, and operating cost assessment. Hub locations determine the required flight range, which affects both flight routes and vehicle design. The flight plan and vehicle performance determine the number of drones needed to meet delivery demand. \rev{Fleet size and vehicle dimensions determine hub area, while flight frequency determines staffing; both affect operating costs.}

\subsection{Delivery System and Demand}\label{sec:definition}

A hub-based delivery system using electric quadrotor drones is considered. Each delivery area is represented by a demand node and assigned to one hub. Hubs provide space for cargo handling, drone deployment, battery charging and replacement, and maintenance. Drones serve demand nodes within their assigned zones. Shipments between different zones are transported through inter-hub connections.

A drone may visit several demand nodes during a flight but must return to its departure hub. Each closed-loop route is counted as one flight. A shipment may require multiple flights to reach its final destination. Therefore, the number of deliveries is distinguished from the number of flights. Each delivery is counted once, regardless of the number of flights needed to complete it.

Delivery demand is represented by origin–destination (OD) shipment counts between demand nodes. The original ground-delivery data are scaled to obtain the planned drone delivery demand. The aggregate OD demand for the monthly planning period is assumed to be known a priori. Demand forecasting and real-time dispatching are not considered.

\subsection{Framework Overview}\label{sec:framework_flow}

The overall framework is shown in Fig.~\ref{fig:framework}. For a specified number of hubs, the locations of delivery demand nodes and candidate hub sites are used to select hub locations and assign demand nodes to service zones. The resulting network determines the required flight range, $R_{\mathrm{req}}$.

The required range is used for both route planning and vehicle sizing. In route planning, candidate flight routes are generated, and their frequencies are optimized to meet the planned OD demand with the minimum number of flights. In vehicle sizing, the motor–propeller combination and battery mass are selected to minimize take-off mass while satisfying the required range and other mission requirements.

The selected vehicle design is used at all hubs within each scenario. However, the number of drones at each hub depends on its flight activity. Flight distances and frequencies are combined with the vehicle's cruise speed and power to calculate flight hours and energy consumption. These results are used to estimate fleet size and battery and motor replacement requirements.

\rev{Hub area is calculated from fleet size and vehicle dimensions, while staffing is estimated from flight frequency and ground-task time.} Rent, wages, battery charging, and component replacement costs are then combined to calculate monthly operating cost and operating cost per delivery.

Table~\ref{tab:stages} summarizes the objectives and conditions of each stage. The design stages are optimized separately, and their outputs are used in subsequent calculations. The framework therefore provides an integrated cost assessment without jointly optimizing the network, flight routes, and vehicle design.

\input{tables_ver3/stages_table.tex}

\section{Hub Location and Flight-Route Planning}\label{sec:network}
\subsection{Hub Location and Required Range}
The range needed to serve essential network connections is used as a vehicle-sizing requirement. The hub-location stage therefore seeks a configuration that reduces this requirement.

The $p$-median facility-location model minimizes total assignment distance, whereas $k$-means clustering minimizes the sum of squared distances to cluster centers. For the hub-location comparison, let $d_{ij}$ denote the distance from demand node $i$ to hub $j$, and let $y_{ij}\in\{0,1\}$ indicate whether node $i$ is assigned to that hub. Their distance-based objectives can be written as
\begin{align}
 \min\quad &\sum_i\sum_j d_{ij}y_{ij},\label{eq:assignment_sum}\\
 \min\quad &\sum_i\sum_j d_{ij}^{2}y_{ij}.\label{eq:assignment_squared}
\end{align}

These objectives improve average distance efficiency but do not directly minimize the longest connection. A small average distance can therefore coexist with a long range requirement for serving some nodes. This is particularly important for quadrotor drones, which have limited flight range due to their relatively low cruise efficiency. A single long connection may require a larger battery and a heavier vehicle, even when the average distance is short. Therefore, the $p$-center principle of minimizing the maximum assignment distance is adopted:

\begin{equation}
 \min\quad\max_i\left\{\sum_j d_{ij}y_{ij}\right\}.\label{eq:assignment_max}
\end{equation}

The conventional $p$-center criterion considers only distances between demand nodes and their assigned hubs and does not directly control inter-hub distances. In the delivery system considered here, local service uses closed-loop flights and shipments between zones pass through hubs. Both local service distances and inter-hub connections must therefore enter the range requirement. An extended $p$-center model is used to evaluate the longest hub--node and inter-hub distances together.

Let $\mathcal N$ be the demand-node set and $\mathcal C_{\mathrm{hub}}\subseteq\mathcal N$ the candidate hub set. Candidates are restricted to representative points of demand areas, and $\mathcal H$ denotes a set of $H$ selected hubs. Distances between representative points are used for inter-area connections, while intra-area delivery distances are included separately when evaluating routes. Each node is assigned to its nearest hub. The nodes assigned to hub $j$ form its service zone $\mathcal S_j(\mathcal H)$. Including all hub pairs for inter-hub transportation, the maximum link distance is
\begin{equation}
 Z(\mathcal H)=\max\left\{
 \max_{i\notin\mathcal H}\min_{j\in\mathcal H}d_{ij},
 \max_{j,\ell\in\mathcal H,\,j\ne\ell}d_{j\ell}
 \right\}.\label{eq:hub}
\end{equation}
The inter-hub term is omitted when there is only one hub.

The hub configuration minimizing the maximum link distance need not be unique. The mean hub-assignment distance over all demand nodes is introduced as a secondary criterion to distinguish hub configurations with the same minimum maximum distance:

\begin{equation}
 \bar D(\mathcal H)=\frac{1}{|\mathcal N|}\sum_{j\in\mathcal H}
 \sum_{i\in\mathcal S_j(\mathcal H)}d_{ij}.\label{eq:assignment_mean}
\end{equation}
Combining the distance criteria and the candidate-site and hub-count constraints yields
\begin{equation}
\begin{aligned}
 \min_{\mathcal H}\quad &Z(\mathcal H)+\varepsilon\bar D(\mathcal H)\\
 \text{s.t.}\quad &\mathcal H\subseteq\mathcal C_{\mathrm{hub}},\\
 &|\mathcal H|=H.
\end{aligned}\label{eq:hub_lex}
\end{equation}

The weight $\varepsilon$ is chosen to be small enough that minimizing the maximum distance takes priority over minimizing the average distance. All possible hub combinations are evaluated for each specified number of hubs.

The selected configuration reflects the spatial trade-off between local service and inter-zone connections. For the selected hub set $\mathcal H^\ast$, let $Z^\ast=Z(\mathcal H^\ast)$. The range requirement passed to vehicle sizing is
\begin{equation}R_{\mathrm{req}}=2Z^\ast.\label{eq:requiredrange}\end{equation}

No additional distance margin is applied to $R_{req}$. A separate hover-time energy reserve is included during vehicle sizing.

\begin{figure}[!t]
    \centering
    \includegraphics[width=0.85\linewidth]{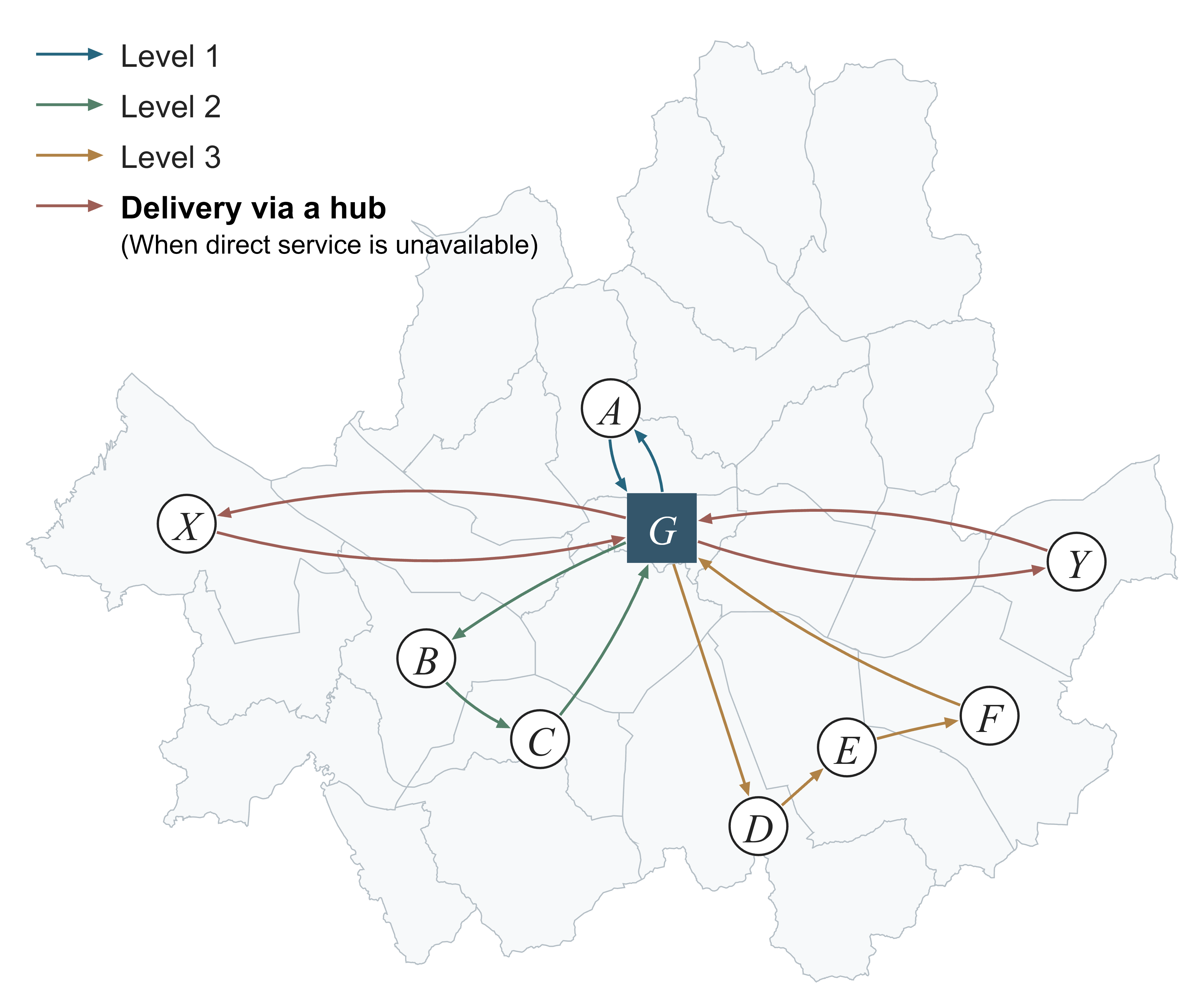}
    \caption{Hub-based single- and multi-stop closed-loop flights.}
    \label{fig:route_examples}
\end{figure}

\subsection{Route Generation and Demand Transformation}
After hub locations and service zones are determined, candidate delivery routes are generated using depth-first search (DFS).

Starting from each hub, DFS extends a visit sequence by adding adjacent nodes one at a time. When a branch cannot be extended, the search backtracks to explore another adjacent node. Further expansion stops when the cumulative distance exceeds the allowed limit. At each step, a direct return to the departure hub is also checked. A closed-loop route is retained if its total length $L_r$, including the return segment, satisfies
\begin{equation}L_r\le R_{\mathrm{lim}}.\label{eq:routecap}\end{equation}
Here, $R_{\mathrm{lim}}$ is the maximum allowed flight distance for each route.

Node connections follow the service-zone assignments and hub-connection rules. Direct delivery between non-hub nodes in different service zones is not allowed. Instead, shipments must pass through their assigned hubs. After reaching a hub other than the departure hub, the route is not extended through another hub in succession. To avoid unnecessary repeated visits, a previously visited node cannot be added again, except for a self-loop at the current node. Thus, revisits such as $A\to B\to A$ are excluded, but the final return to the departure hub is allowed. Delivery within a demand area is represented by a self-loop at its node. A sequence such as $A\to A\to B$ is allowed, whereas three or more consecutive visits to the same node are excluded.

Routes can be classified by the number of visits. In Fig.~\ref{fig:route_examples}, $G$ denotes the hub. A level-1 route, such as $G\to A\to G$, visits one node. A level-2 route, such as $G\to B\to C\to G$, visits two nodes in sequence. Only routes satisfying the distance and search rules are retained. Demand such as $X\to Y$ that cannot be served directly within the flight-distance limit, including the return to the hub, is split into segments through hub $G$.
% \begin{figure}[!t]
% \centering
% \includegraphics[width=0.95\linewidth]{image_ver3/Route_examples_Figure2_EN_KO4.png}
% \caption{Hub-based single- and multi-stop closed-loop flights.}
% \label{fig:route_examples}
% \end{figure}
% This procedure constructs the candidate route set $\mathcal R$ under the specified search rules and maximum depth.

Planned OD demand is obtained by scaling the original ground delivery shipment counts. With $q_{od}$ denoting the original shipment count for OD pair $od$ and $\beta$ the demand scaling factor, the resulting demand is given by
\begin{equation}
 k_{od}=\lfloor\beta q_{od}\rfloor,\quad K=\sum_{od}k_{od}.\label{eq:demand}
\end{equation}
Rounding down each OD entry means that $K$ may differ from $\beta$ times the total original demand. Here, $k_{od}$ is the planned OD demand for pair $od$, and $K$ is its total across OD pairs, measured in deliveries. The same factor is applied to every OD pair, preserving spatial OD proportions apart from rounding.
\begin{figure}[!t]
\centering
\subfloat[Decomposition of inter-zone demand]{\includegraphics[width=\linewidth]{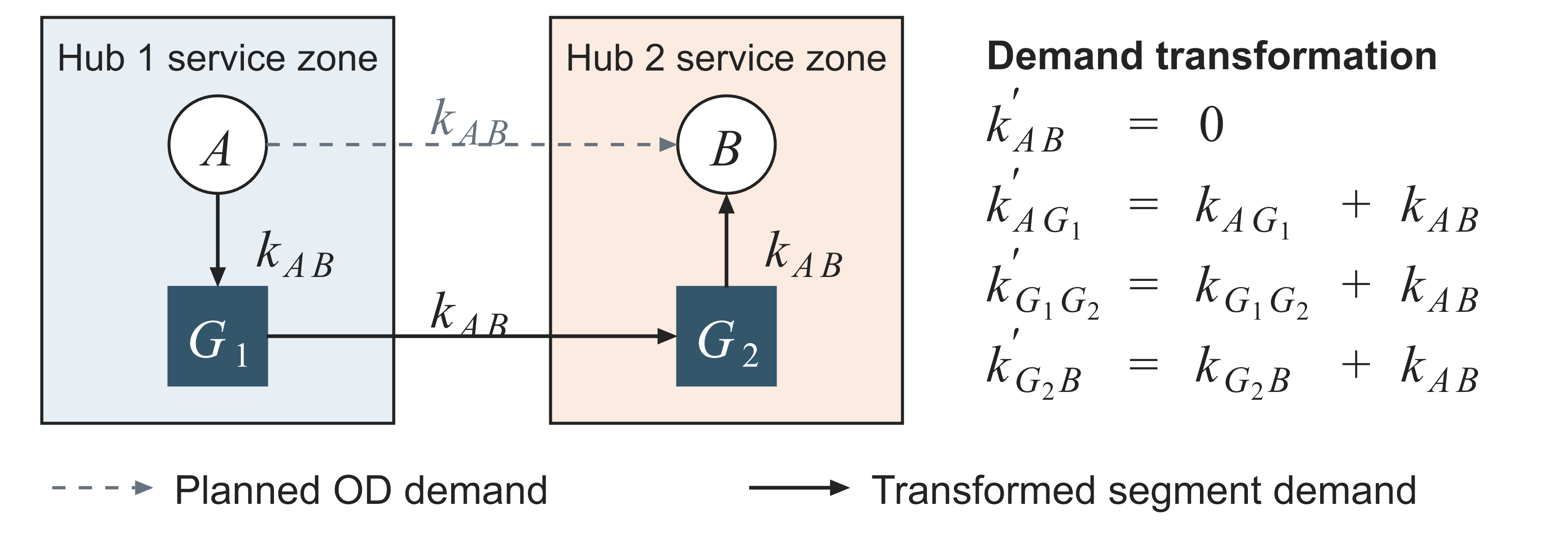}}\\[0.8ex]
\subfloat[Intra-zone demand when direct service is unavailable]{\includegraphics[width=\linewidth]{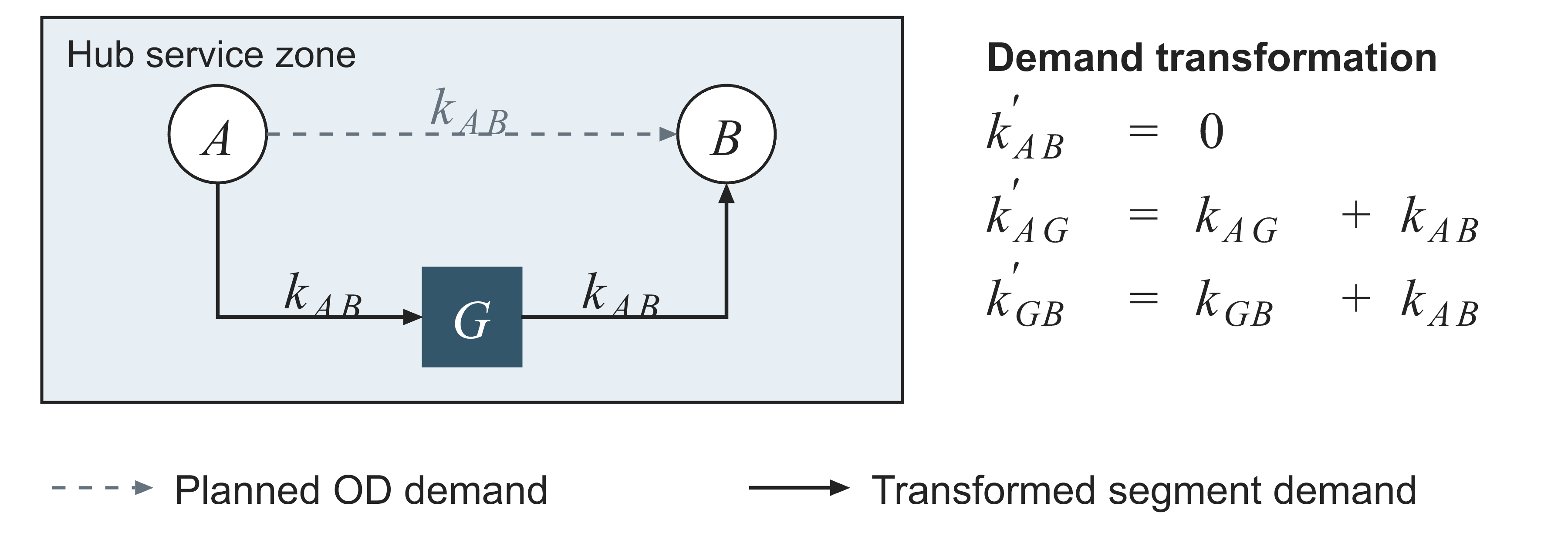}}
\caption{Decomposition of planned OD demand through hubs.}
\label{fig:demand_decomposition}
\end{figure}

To plan operations over the candidate routes, planned OD demand is transformed into segment demand on the service segments used by those routes. Fig.~\ref{fig:demand_decomposition} distinguishes shipments between different hub zones from shipments within a zone that cannot be served directly. Dashed arrows represent planned OD demand before decomposition, and solid arrows represent the segments to which that demand is reassigned.

In Fig.~\ref{fig:demand_decomposition}(a), demand $k_{AB}$ between non-hub nodes $A$ and $B$, assigned to $G_1$ and $G_2$, respectively, is split into three segments: $A\to G_1$, $G_1\to G_2$, and $G_2\to B$. Even when two nodes belong to the same hub $G$, no feasible candidate flight may include their direct OD segment. Their demand is then split into $A\to G$ and $G\to B$, as shown in Fig.~\ref{fig:demand_decomposition}(b).

In both cases, the planned OD demand $k_{AB}$ is added to the existing demand on each resulting segment, and the direct-segment demand becomes $k'_{AB}=0$. The update equations in the figure illustrate the transformation of demand for one OD pair. The resulting segments may be served by different flights.
% \begin{figure}[!t]
% \centering
% \subfloat[Decomposition of inter-zone demand]{\includegraphics[width=\linewidth]{image_ver3/Demand_decomposition_Figure3a_EN_KO4.png}}\\[0.8ex]
% \subfloat[Intra-zone demand when direct service is unavailable]{\includegraphics[width=\linewidth]{image_ver3/Demand_decomposition_Figure3b_EN_KO4.png}}
% \caption{Decomposition of planned OD demand through hubs.}
% \label{fig:demand_decomposition}
% \end{figure}

Demand decomposition leaves the planned delivery count $K$ unchanged, while the total segment demand $K'=\sum_e k'_e$ may increase.

\subsection{Flight-Frequency Optimization}
The combination and frequency of candidate routes are selected to serve all segment demand with as few flights as possible. This is formulated as a mixed-integer linear program (MILP) for route-specific flight counts.

As defined earlier, segment demand is the number of shipments to be transported along each service segment. Each execution of a route provides capacity for one such shipment on each included segment.

Let the integer $x_r$ denote the monthly number of flights on candidate route $r$. These values form the route-frequency vector $\mathbf{x}$ of length $|\mathcal R|$; $x_r=0$ indicates that a route is unused. The set $\mathcal E$ contains directed service segments between nodes; self-loops represent deliveries within a demand area. In the route--segment incidence matrix $\mathbf{R}\in\{0,1\}^{|\mathcal R|\times|\mathcal E|}$, rows correspond to candidate routes and columns to these segments. Entry $R_{re}$ is one if route $r$ contains segment $e$ and zero otherwise.

\begin{figure}[!t]
\centering
\includegraphics[width=0.95\linewidth]{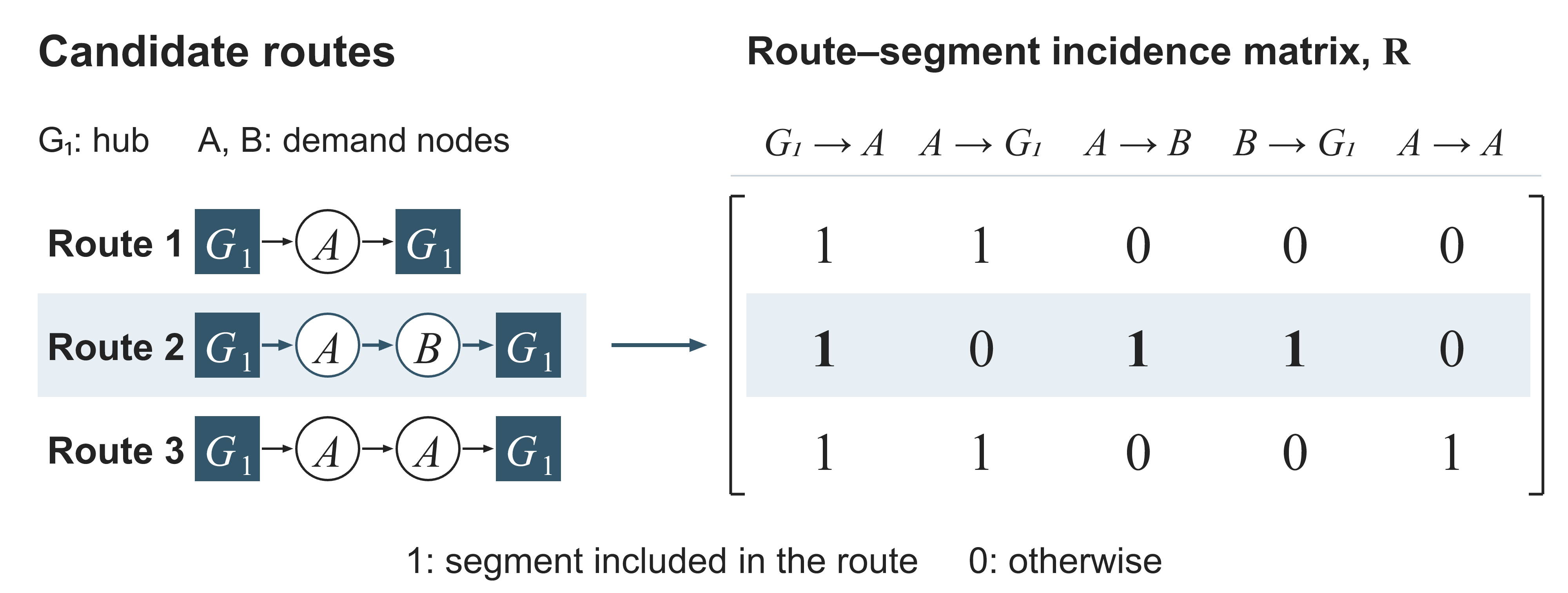}
\caption{Illustrative construction of the route--segment incidence matrix $\mathbf{R}$.}
\label{fig:route_incidence}
\end{figure}

Fig.~\ref{fig:route_incidence} illustrates $\mathbf{R}$ using three routes and five segments in a network where hub $G_1$ serves demand nodes $A$ and $B$. The highlighted route 2 contains $G_1\to A$, $A\to B$, and $B\to G_1$, so its row is $(1,0,1,1,0)$ in the displayed column order. The self-loop $A\to A$ in route 3 represents delivery within the area associated with node $A$ and is treated like any other service segment.

With the transformed demand vector $\mathbf{k}'$ containing $k'_e$, the flight plan is obtained from
\begin{equation}
 \begin{aligned}
 \min_{\mathbf{x}}\quad &F=\mathbf{1}^{\mathsf T}\mathbf{x}\\
 \text{s.t.}\quad &(\mathbf{R}^{\mathsf T}\mathbf{x})_e\ge k'_e,\qquad \forall e\in\mathcal E,\\
 &\mathbf{x}\in\mathbb Z_{\ge0}^{|\mathcal R|}.
 \end{aligned}\label{eq:flightmilp}
\end{equation}
The vector $\mathbf{1}$ has $|\mathcal R|$ unit entries, and $(\mathbf{R}^{\mathsf T}\mathbf{x})_e$ is the total number of service units available on segment $e$. Available service can exceed demand on some segments, but the excess is not counted as additional delivery demand. This stage minimizes flight count; distance, energy, and operating cost are subsequently evaluated for the selected flight plan.

\section{Mission-Based Drone Sizing and Performance Evaluation}\label{sec:vehicle}

Vehicle design is formulated as a constrained optimization problem to minimize take-off mass while satisfying the range requirement determined by the delivery network and other mission conditions. Vehicle mass and flight performance are evaluated using the sizing and propulsion models described in the following subsections.

\subsection{Vehicle Sizing and Mass Estimation}

The mission inputs are required range, payload mass, hover reserve time, and cruise altitude. For a given motor–propeller combination $c$ and battery mass $m_b$, the sizing model calculates vehicle dimensions, component masses, and total take-off mass. Battery voltage is held constant across all designs, while battery capacity is determined through the sizing process.

\begin{figure}[!t]
\centering
\includegraphics[width=\linewidth]{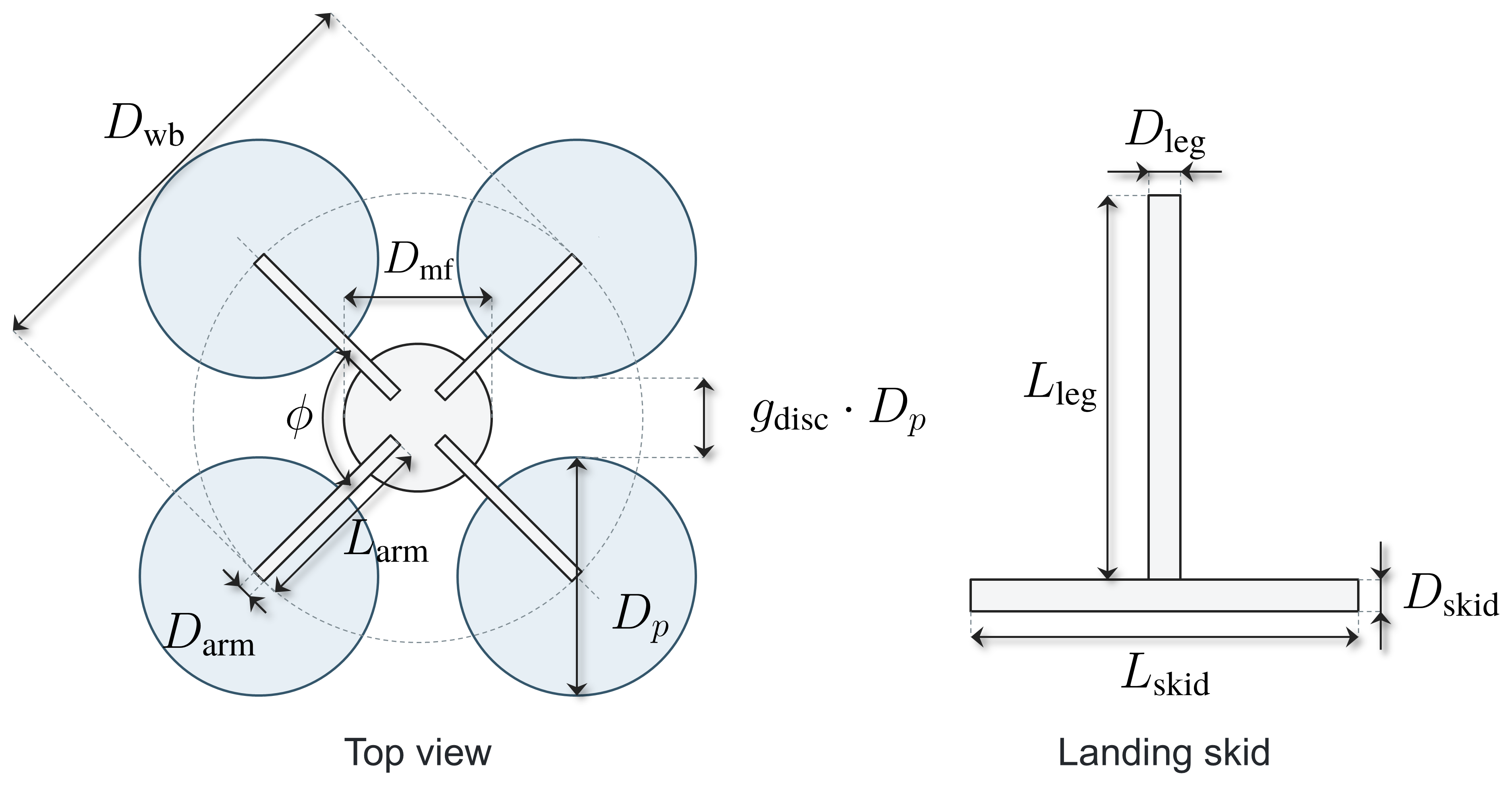}
\caption{Representative quadcopter geometry and dimensions.}
\label{fig:vehicle_geometry}
\end{figure}

The vehicle is an electric quadcopter. The left panel of Fig.~\ref{fig:vehicle_geometry} shows the rotor and main-frame layout in top view, while the right panel shows representative landing-skid dimensions. Here, $D_p$ is propeller diameter, $D_{\mathrm{wb}}$ is the distance between opposite motor centers (wheelbase), and $D_{\mathrm{mf}}$ is main-frame diameter. The rotor-arm length is $L_{\mathrm{arm}}$, with $D_{\mathrm{arm}}$ denoting its outer diameter. For the landing gear, $L_{\mathrm{leg}}$ denotes leg length and $D_{\mathrm{leg}}$ denotes leg outer diameter. The skid length is $L_{\mathrm{skid}}$, with $D_{\mathrm{skid}}$ denoting its outer diameter.

Manufacturer specifications provide the candidate motor mass $m_m$ and propeller mass $m_p$\cite{tmotor,apc}. Structural and auxiliary component masses are estimated using the preliminary-sizing relationships of Jeong\cite{jeong2021}, which combine geometric calculations with empirical models. The design take-off mass is
\begin{equation}
m_{\mathrm{TO}}=4(m_m+m_p)+m_b+m_{\mathrm{af}}+m_{\mathrm{av}}+m_{\mathrm{pl}}+m_{\mathrm{misc}}.
\label{eq:mass}
\end{equation}

Here, the subscripts $m$, $p$, $b$, $af$, $av$, $pl$, and $misc$ denote motor, propeller, battery, airframe, avionics, payload, and miscellaneous components, respectively. The factor of four accounts for the four motors and propellers.

The airframe mass is
\begin{equation}
m_{\mathrm{af}}=m_{\mathrm{mf}}+4(m_{\mathrm{arm}}+m_{\mathrm{mm}})+m_{\mathrm{ls}}+m_{\mathrm{bm}}+m_{\mathrm{ph}},
\label{eq:mass_airframe}
\end{equation}
where the subscripts $\mathrm{mf}$, $\mathrm{arm}$, $\mathrm{mm}$, $\mathrm{ls}$, $\mathrm{bm}$, and $\mathrm{ph}$ refer to the main frame, rotor arm, motor mount, landing skids, battery mount, and pipe holders. Four rotor arms and four motor mounts are included.

The main dimensions are calculated based on the selected propeller diameter. With a clearance of $g_{\mathrm{disc}}D_p$ between adjacent propellers and an angle $\phi$ between rotor arms,
\begin{equation}
\begin{gathered}
D_{\mathrm{wb}}=\frac{(1+g_{\mathrm{disc}})D_p}{\sin(\phi/2)},\\
D_{\mathrm{mf}}=r_{\mathrm{mf}}D_{\mathrm{wb}},\quad L_{\mathrm{arm}}=r_{\mathrm{al}}D_{\mathrm{wb}}.
\end{gathered}\label{eq:geometry_sizing}
\end{equation}

The ratios $r_{\mathrm{mf}}$ and $r_{\mathrm{al}}$ relate main-frame diameter and rotor-arm length to wheelbase. Propeller selection therefore affects propulsion performance and structural mass, as well as the vehicle dimensions used to estimate the required space at delivery hub facilities.

The main frame and mounts are approximated as plates, whereas the rotor arms and landing skids are modeled as hollow tubes. For these geometries, component masses depend on dimensions, thickness, and material density, with a mass-reduction factor applied to plate components. Rotor-arm and landing-gear dimensions scale with propeller size. Mounting structures and connectors are sized using motor specifications, battery specifications, and empirical relationships. When battery mass changes, stored energy is recalculated along with the masses of the battery mount and associated components.

Avionics are assigned a fixed mass. Miscellaneous components comprise wiring and structural accessories:
\begin{equation}
m_{\mathrm{misc}}=k_{\mathrm{wire}}\bigl[4(m_m+m_p)+m_b\bigr]+k_{\mathrm{subs}}m_{\mathrm{af}}.
\label{eq:mass_misc}
\end{equation}
Here, $k_{\mathrm{wire}}$ is the wiring-to-propulsion-system mass ratio, and $k_{\mathrm{subs}}$ is the accessory-to-airframe mass ratio. Table~\ref{tab:mass_constants} lists the design constants and fixed assumptions. Detailed component dimensions and mass relationships are provided in Chapters 4 and 5 of Jeong\cite{jeong2021}.
\input{tables_ver3/mass_constants_table.tex}

\subsection{Propulsion Performance and Range Evaluation}

% Good
Mission feasibility is evaluated by first calculating the thrust and power required for hover, then determining forward-flight speed, power, and battery-limited range. A propulsion operating point comprises motor rotational speed and the corresponding thrust, torque, motor current, electrical power, and airspeed. These quantities are calculated using the coupled propulsion and force-balance models, with air density determined by the cruise altitude.

Manufacturer-supplied static propeller performance data at different rotational speeds provide a common basis for preliminary sizing across many commercial propulsion candidates\cite{apc}. These data are combined with motor specifications and a propulsion model to calculate thrust, torque, and efficiency\cite{nam2024,tmotor}. The thrust $T$ and torque $Q$ of one propeller are
\begin{equation}
T=C_T\rho n^2D_p^4,\quad Q=\frac{C_P}{2\pi}\rho n^2D_p^5,
\label{eq:prop_loads}
\end{equation}
where $n$ [rev/s] is rotational speed, $\rho$ is air density, and $C_T$ and $C_P$ are the static thrust and power coefficients. Hover is evaluated without external forward inflow. The rotational speed and torque must satisfy $T=m_{\mathrm{TO}}g/4$, assuming equal loading of the four rotors, where $g$ is gravitational acceleration.
\begin{figure}[!t]
\centering
\includegraphics[width=0.8\linewidth]{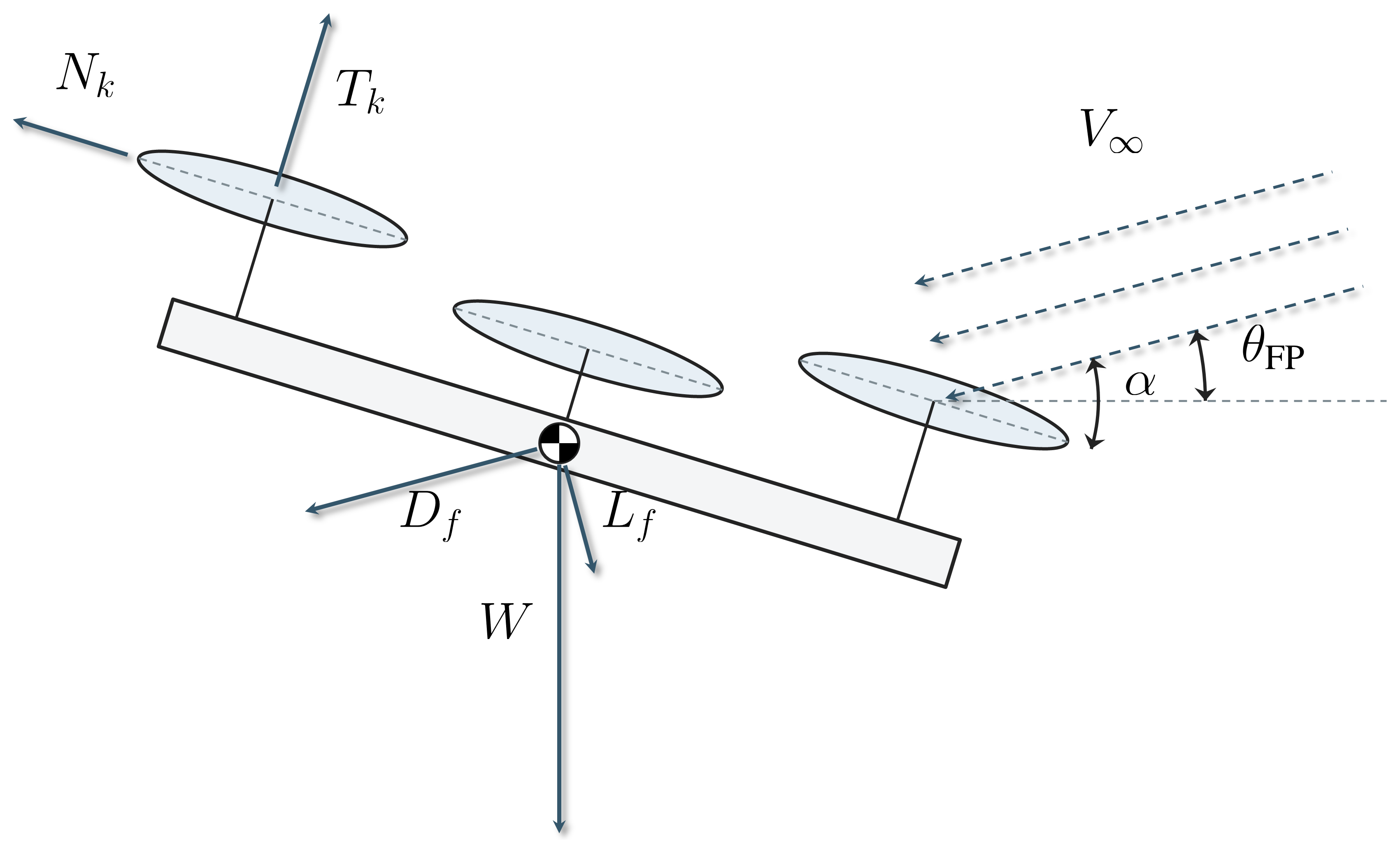}
\caption{Force and attitude definitions in forward flight.}
\label{fig:vehicle_force_balance}
\end{figure}

The electrical power drawn from the battery equals the total rotor shaft power plus the power losses in the motor--ESC units. With angular speed $\omega=2\pi n$ and loss power $P_L$ per motor--ESC unit, the combined efficiency $\eta_m$ and total battery-side power are
\begin{equation}
\eta_m=\frac{Q\omega}{Q\omega+P_L},\quad P_{\mathrm{batt}}=\frac{4Q\omega}{\eta_m}.
\label{eq:battery_power}
\end{equation}

Losses are approximated by an equivalent-circuit model using motor speed constant, no-load current, internal resistance, and battery voltage, with partial throttle represented by a duty ratio. Efficiency is reevaluated at each operating point to calculate hover and forward-flight power, $P_{\mathrm{hv}}$ and $P_{\mathrm{ff}}$. Further model details are given by Nam et al.\cite{nam2024} and Saemi and Benedict\cite{saemi2023flight}.

Static coefficients are also used to approximate forward-flight propulsion, omitting the dependence on advance ratio and inflow angle considered by Pollet et al.\cite{pollet2021}. Vehicle drag and attitude enter through the force balance in Fig.~\ref{fig:vehicle_force_balance}. Here, $T_k$ and $N_k$ are the thrust and in-plane force of rotor $k$, $D_f$ and $L_f$ are vehicle drag and the downward vertical aerodynamic force, and $W=m_{\mathrm{TO}}g$ is weight. The flight-path angle relative to the horizontal is $\theta_{\mathrm{FP}}$. For steady level flight, equal rotor thrust ($T_k=T$), and negligible in-plane force ($N_k=0$), the horizontal and vertical balances become
\begin{equation}
\begin{gathered}
4T\sin\alpha=D_f=\tfrac12\rho V^2 S_{\mathrm{eff}} C_D,\\
4T\cos\alpha=W+L_f=m_{\mathrm{TO}}g+\tfrac12\rho V^2 S_{\mathrm{eff}} C_L.
\end{gathered}\label{eq:balance}
\end{equation}
Here, $V$ is airspeed and $\alpha$ is the angle between the thrust axis and the vertical. This study assumes still air and uses the airspeed $V$ to represent the magnitude of the inflow velocity $V_\infty$ shown in Fig.~\ref{fig:vehicle_force_balance}.
% \begin{figure}[!t]
% \centering
% \includegraphics[width=0.8\linewidth]{image_ver3/Drone_force_KO4.png}
% \caption{Force and attitude definitions in forward flight.}
% \label{fig:vehicle_force_balance}
% \end{figure}

The effective area $S_{\mathrm{eff}}$ combines the top and frontal projected areas according to attitude. Geometric similarity is used to scale the reference areas with mass, and the payload frontal area is added:
\begin{equation}
\begin{gathered}
S_{\mathrm{eff}}=S_{\mathrm{top}}\sin\alpha+S_{\mathrm{front}}\cos\alpha,\\
S_{\mathrm{top}}=S_{\mathrm{top,ref}}\left(\frac{m_{\mathrm{TO}}}{m_{\mathrm{ref}}}\right)^{2/3},\\
S_{\mathrm{front}}=S_{\mathrm{front,ref}}\left(\frac{m_{\mathrm{TO}}}{m_{\mathrm{ref}}}\right)^{2/3}+S_{\mathrm{front,box}}.
\end{gathered}\label{eq:effective_area}
\end{equation}

The reference vehicle is characterized by its mass $m_{\mathrm{ref}}$ and projected areas $S_{\mathrm{top,ref}}$ and $S_{\mathrm{front,ref}}$; $S_{\mathrm{front,box}}$ denotes the payload frontal area. Table~\ref{tab:aero_constants} gives the aerodynamic and geometric inputs. At each propulsion operating point, \eqref{eq:balance} is solved together with \eqref{eq:effective_area} to obtain $\alpha$ and $V$; $P_{\mathrm{ff}}$ is then calculated from \eqref{eq:battery_power}. The general force balance and aerodynamic model are described by Pollet et al.\cite{pollet2021}.
\input{tables_ver3/aero_constants_table.tex}

Range is evaluated from speed, power, and the energy remaining after allocating the hover reserve. For battery specific energy $e_b$ [Wh/kg] and hover reserve time $t_{\mathrm{hv}}$,
\begin{equation}
E_b=3600e_bm_b,\quad E_{\mathrm{hv}}=P_{\mathrm{hv}}t_{\mathrm{hv}},\quad E_{\mathrm{ff}}=E_b-E_{\mathrm{hv}}.
\label{eq:energy}
\end{equation}
The terms are battery energy, hover reserve energy, and energy available for forward flight, respectively. The energy available for forward flight must be positive after accounting for the hover reserve. Forward-flight power is limited to 90\% of the combined maximum motor power, $P_{\mathrm{hv}}<P_{\mathrm{ff}}\le0.9\times4P_{m,\max}$.

% LHT - Need to be revised extensively
\subsection{Vehicle Design Optimization}

For a given motor–propeller combination $c$ and battery mass $m_b$, the maximum range is determined by evaluating the feasible forward-flight operating points. Let $\mathcal{U}(c,m_b)$ denote the set of operating points satisfying the conditions described in Section IV-B. The maximum range is

\begin{equation}
R_{\max}(c,m_b)=\max_{u\in\mathcal U(c,m_b)}\frac{V(u)[E_b-E_{\mathrm{hv}}]}{1000P_{\mathrm{ff}}(u)},
\label{eq:range}
\end{equation}

where range is in kilometers. The speed and power at this range-maximizing operating point are used together as cruise speed $V_{\mathrm{cr}}$ and cruise power $P_{\mathrm{cr}}$, respectively.

The vehicle design optimization minimizes take-off mass while satisfying the range requirement determined by the delivery network. The motor–propeller combination is selected from a discrete component database, while battery mass is treated as a continuous design variable.

\begin{figure}[!t]
\centering
\includegraphics[width=0.9\linewidth]{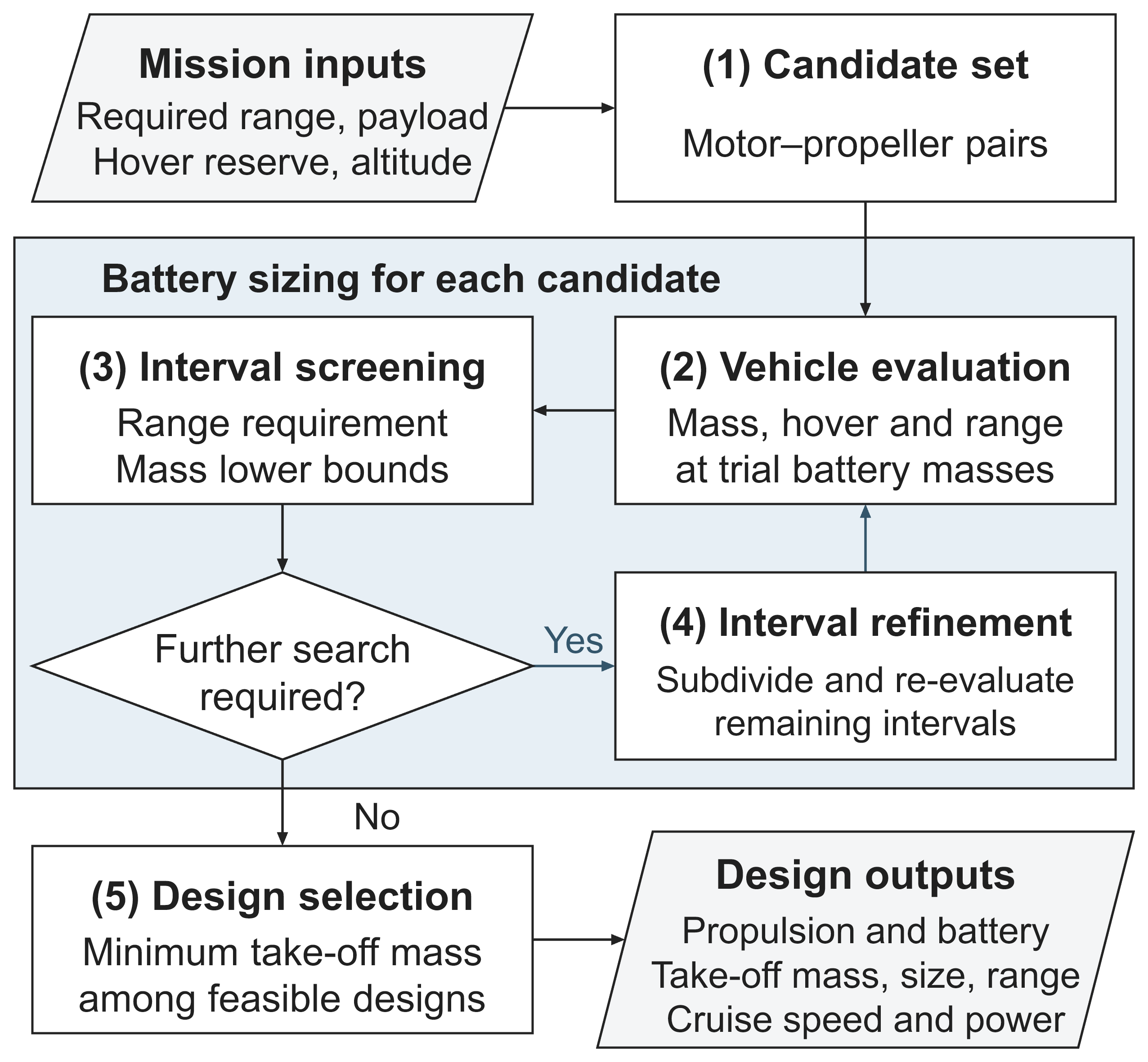}
\caption{Vehicle design search for minimum take-off mass.}
\label{fig:sizing}
\end{figure}

Increasing battery mass increases stored energy but also increases vehicle weight and the power required for flight. Consequently, the achievable range does not necessarily increase with battery mass. Changes in battery mass also affect structural mass, hover power, and forward-flight performance, which are recalculated during optimization.

Because motor–propeller combinations are discrete, a systematic search is used to solve the design optimization problem, as illustrated in Fig.~\ref{fig:sizing}. The database contains 303 T-Motor motors and 428 APC propellers, giving 129,684 possible combinations\cite{tmotor,apc}. The same component database and physical models are used in every hub-count scenario to examine the effects of different mission requirements.

Let $\mathcal C_{\mathrm{prop}}$ be the candidate motor--propeller set. Configuration $c\in\mathcal C_{\mathrm{prop}}$ is discrete, whereas battery mass $m_b$ is continuous. For the required range of each network,

\begin{equation}
\begin{aligned}
\min_{c,m_b}\quad&m_{\mathrm{TO}}(c,m_b)\\
\text{s.t.}\quad&R_{\max}(c,m_b)\ge R_{\mathrm{req}},\\
&m_b\in[0.05,20]\ \mathrm{kg},\quad c\in\mathcal C_{\mathrm{prop}}.
\end{aligned}\label{eq:design}
\end{equation}
The payload mass, hover reserve time, and cruise altitude remain fixed during each design search.

For a fixed motor–propeller combination, take-off mass increases with battery mass. Therefore, the take-off mass at the lower end of each battery-mass interval provides a lower bound for that interval. Intervals whose lower bounds exceed the lowest feasible take-off mass found so far can be excluded from further evaluation.

% \begin{figure}[!t]
% \centering
% \includegraphics[width=0.95\linewidth]{image_ver3/Drone_sizing_Figure7_EN_KO4.png}
% \caption{Vehicle design search for minimum take-off mass.}
% \label{fig:sizing}
% \end{figure}

The remaining battery-mass intervals are subdivided and reevaluated to identify the minimum-mass feasible design for each motor–propeller combination. The lightest feasible design among all combinations is then selected. Its cruise speed and power are used in the operational and cost assessment in Section~\ref{sec:cost}.

\subsection{Evaluation Against Commercial Drone Specifications}

The vehicle sizing and performance models are evaluated by comparing their predictions with commercial drone specifications. For each commercial drone, the closest available motor and propeller are selected from the T-Motor and APC databases, respectively, while the battery energy is matched to that of the commercial vehicle. With zero additional payload, the models calculate take-off mass, wheelbase, and hover endurance. The corresponding propulsion operating points, including motor rotational speed and motor–ESC efficiency, are also calculated. No vehicle design optimization is performed in these comparisons. Table~\ref{tab:commercial} presents the calculated results alongside the published specifications of four commercial drones.
\input{tables_ver3/commercial_table.tex}

The estimated dimensions are broadly comparable to those of the commercial vehicles. For the M690A and Mavic 3T, the calculated wheelbases are 695 and 367 mm, respectively, close to the published values of 699 and 380 mm. The Matrice 100 and Inspire 1 wheelbase estimates are smaller than the published values. Estimated masses and endurance values are also broadly comparable, with some predictions above and others below the reported specifications.

The calculated endurance represents hovering, whereas the published endurance values reflect product-specific flight and payload conditions. Differences in propulsion components, geometry, and accessories also limit direct comparison. Nevertheless, the results suggest that the geometric and mass-estimation assumptions in Table~\ref{tab:mass_constants} are reasonable for preliminary vehicle sizing.

\section{Operational Requirements and Cost Assessment}\label{sec:cost}

The flight plans obtained in Section~\ref{sec:network} and the vehicle characteristics determined in Section~\ref{sec:vehicle} provide the basis for estimating operating requirements and costs. Flight activity and vehicle performance are used to calculate energy consumption, component replacement, and fleet requirements. \rev{Fleet size and vehicle dimensions are then used to estimate hub space, while flight frequency and ground-task time determine staffing.}

Hub configurations and vehicle designs are compared using the same cost scope and unit prices. Monthly operating cost includes charging electricity and component replacement, as well as hub operating expenses. Initial vehicle acquisition cost is reported separately as an indicator of investment requirements, without amortization into monthly operating cost. All monetary inputs and costs are expressed in USD, using the fixed conversion of 1 USD = 1,400 KRW for source prices quoted in Korean won.

The following calculations apply to a scenario with $H$ hubs. Subscript $j$ denotes an individual hub within that scenario; scenario totals are obtained by summing the hub-level quantities.

\subsection{Flight Time, Energy, and Fleet Requirements}
Operating cost assessment first requires translating the flight plan into hub-level flight activity and fleet requirements. Route frequencies and distances are aggregated by departure hub, including both service segments and return travel to the departure hub.

Let $\Omega_{rj}=1$ if route $r$ departs from hub $j$, and zero otherwise. Each route has one departure hub, so $\sum_j\Omega_{rj}=1$. With segment-distance vector $\boldsymbol{\updelta}$ and route-distance vector $\mathbf L$, route distances, monthly flight distances, and flight counts are
\begin{equation}
\mathbf L=\mathbf R\boldsymbol{\updelta},\label{eq:route_distance}\\
\end{equation}
\begin{equation}
d_j^{\mathrm{tot}}=\sum_r\Omega_{rj}x_rL_r,\label{eq:hub_distance}\\
\end{equation}
\begin{equation}
\begin{gathered}
F_j=\sum_r\Omega_{rj}x_r,\quad F=\sum_jF_j.\label{eq:hub_flights}
\end{gathered}
\end{equation}
Here, $d_j^{\mathrm{tot}}$ and $F_j$ are the monthly flight distance and flight count at hub $j$, respectively; distances are calculated in meters.

Cruise performance converts flight distance into flight time and energy. Fleet size is then estimated from the available monthly operating time per vehicle. Let $n_{\mathrm{day}}$ be operating days per month and $h_{\mathrm{day}}$ effective daily operating hours. Hub-level flight time $T_j$, energy consumption $E_j$, and fleet size $N_j$ are
\begin{equation}
\begin{gathered}
T_j=d_j^{\mathrm{tot}}/V_{\mathrm{cr}},\quad E_j=P_{\mathrm{cr}}T_j/(3.6\times10^6),\\
N_j=\left\lceil\frac{T_j}{n_{\mathrm{day}}h_{\mathrm{day}}3600}\right\rceil.
\end{gathered}\label{eq:fleet}
\end{equation}
The units are seconds for $T_j$, meters per second for $V_{\mathrm{cr}}$, watts for $P_{\mathrm{cr}}$, kilowatt-hours for $E_j$, and hours per day for $h_{\mathrm{day}}$. Total fleet size is $\sum_jN_j$, with rounding applied separately at each hub. Consequently, the distribution of flight activity across hubs can affect the spare capacity implied by integer fleet allocation even when total activity is unchanged.

\subsection{Battery and Motor Replacement}
Monthly replacement quantities approximate component use under repeated operation. Battery replacement is based on cumulative energy consumption in equivalent full cycles; motor replacement is based on cumulative operating hours for four motors per vehicle.

With battery energy expressed as $B=E_b/(3.6\times10^6)$ [kWh], the monthly replacement quantities at hub $j$ are calculated from battery cycle life $L_b$ [cycles] and motor service life $L_m$ [h] as
\begin{equation}
n_{b,j}=\left\lceil E_j/(L_b B)\right\rceil,\quad n_{m,j}=\left\lceil4T_j/(3600L_m)\right\rceil.
\label{eq:replacement}
\end{equation}
The product $L_bB$ is the energy delivered over a battery's assumed life. Both quantities are rounded up at each hub before costs are aggregated.

\begin{figure}[!t]
\centering
\includegraphics[width=0.85\linewidth]{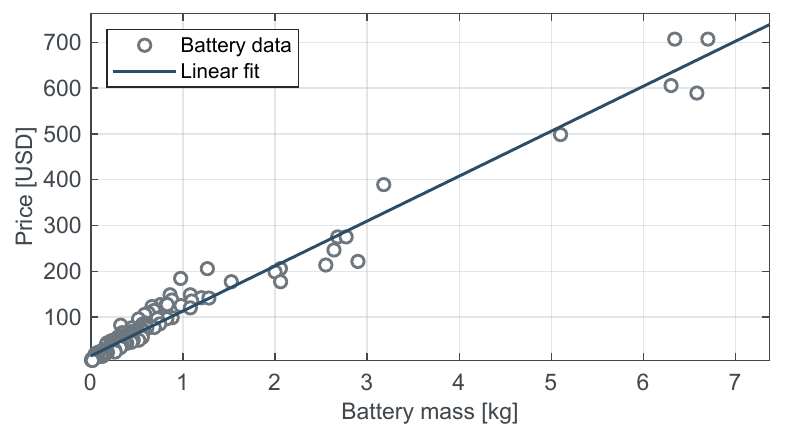}
\caption{Commercial battery mass--price relationship and linear fit.}
\label{fig:battery_price_fit}
\end{figure}

Battery unit price is estimated from publicly listed masses and retail prices of 198 commercial lithium-polymer batteries in the Korean online market. Fig.~\ref{fig:battery_price_fit} shows the observations and least-squares linear fit:
\begin{equation}
p_b(m_b)=98.24\,m_b+14.92\quad[\mathrm{USD}].
\label{eq:battery_price}
\end{equation}
Battery mass is in kilograms. Products with different cell counts, capacities, and discharge ratings are pooled to obtain a mass-based price estimate for comparative assessment.

Motor unit price is retrieved from the collected price data using the selected motor model and speed constant (KV). Vehicle sizing thus affects component prices and replacement cost as well as energy consumption. The assumed lifetimes $L_b$ and $L_m$ are varied in the sensitivity analysis.

\subsection{Hub Space, Staffing, and Operating Cost}

\rev{Hub rent depends on fleet size, vehicle dimensions, and shared fixed space. Staffing combines flight-dependent ground-task personnel with base personnel for hub management and operating support. Let $h_{\mathrm{hub}}$ denote staffed hub operating hours per day and $f_j=F_j/(n_{\mathrm{day}}h_{\mathrm{hub}})$ the average number of flights per operating hour at hub $j$.}

\begin{equation}
    A_j=a_1N_jD_{\mathrm{wb}}^2+a_2,\quad \rev{S_j=\lceil \frac{b_1 f_j}{60}\rceil+b_2}.
    \label{eq:hubcostsize}
\end{equation}

Here, $A_j$ is hub area and $S_j$ is the estimated concurrent staffing requirement at hub $j$, based on the average hourly number of flights. Coefficient $a_1$ represents vehicle placement and operating clearances, while $a_2$ is shared fixed space for cargo handling and support functions. \rev{Coefficient $b_1$ is ground-task labor per flight [person $\cdot$ min/flight], and $b_2$ is the integer number of additional base staff per hub. Task staffing is rounded up separately at each hub before $b_2$ is added.} Vertiport design guidance\cite{faa_vertiport_2022} and neighborhood delivery-hub demonstrations\cite{urbanfreightlab_ndh_2021} provide background for distinguishing operating space and support functions.

\rev{Monthly rent is calculated from hub area, while labor cost is calculated from the person-hours required to maintain concurrent staffing throughout hub operation. Hub-level utilities and maintenance costs are then added:}
\begin{gather}
C_{\mathrm{lease},j}=p_AA_j,\quad \rev{C_{\mathrm{labor},j}=w_LS_jh_{\mathrm{hub}}n_{\mathrm{day}}},\label{eq:rent_labor}\\
C_{\mathrm{hub}}=\sum_j\left(C_{\mathrm{lease},j}+C_{\mathrm{labor},j}+c_u+c_s\right).
\label{eq:hub_cost_breakdown}
\end{gather}
Here, $p_A$ is the monthly rent per unit area, \rev{$w_L$ is the labor cost per staffed person-hour [USD/(person $\cdot$ h)], and $c_u$ and $c_s$ are the monthly utility and maintenance costs per hub, respectively. The product $S_jh_{\mathrm{hub}}n_{\mathrm{day}}$ represents monthly person-hours, rather than the total headcount needed to cover individual shifts.} The hub utility cost $c_u$ excludes electricity used for charging drone batteries. Battery charging is calculated separately as part of the operating cost.

\subsection{Total Operating Cost and Evaluation Metrics}
Unit prices are applied to energy consumption and component replacement quantities, and hub cost is added to obtain monthly scenario cost. Let $p_E$ [USD/kWh] be the electricity price, $p_b$ [USD/battery] the battery price from \eqref{eq:battery_price}, and $p_m$ [USD/motor] the selected motor's unit price. The monthly cost components and their total are
\begin{equation}
C_{\mathrm{charge}}=p_E\sum_jE_j,\label{eq:charge_cost}\\
\end{equation}
\begin{equation}
C_{\mathrm{battery}}=p_b\sum_jn_{b,j},\label{eq:battery_cost}\\
\end{equation}
\begin{equation}
C_{\mathrm{motor}}=p_m\sum_jn_{m,j},\label{eq:motor_cost}\\
\end{equation}
\begin{equation}
C_{\mathrm{total}}=C_{\mathrm{charge}}+C_{\mathrm{battery}}+C_{\mathrm{motor}}+C_{\mathrm{hub}}.
\label{eq:cost_components}
\end{equation}
Each cost term is expressed in USD per month.

Operating cost per delivery, $c_{\mathrm{OD}}$, is the primary metric for comparing hub configurations and vehicle designs. Its denominator $K$ is the total planned OD demand, measured in deliveries. Cost per flight, $c_F$, is a supplementary metric for interpreting flight operations:
\begin{equation}
c_{\mathrm{OD}}=C_{\mathrm{total}}/K,\quad c_F=C_{\mathrm{total}}/F.
\label{eq:metrics}
\end{equation}

\subsection{Initial Vehicle Acquisition Cost}\label{sec:acquisition}
Initial fleet investment is estimated by multiplying unit vehicle price $p_{\mathrm{dr}}$ by total fleet size.
% This unit price is obtained from a regression relating commercial vehicle price to maximum take-off mass (MTOM), using the design take-off mass $m_{\mathrm{TO}}$ as the input:
\rev{A quadratic curve was fitted to maximum take-off mass (MTOM) and price data for five commercial drones. The fit is evaluated at the design take-off mass $m_{\mathrm{TO}}$ in kilograms to estimate the unit price:}
\begin{gather}
p_{\mathrm{dr}}=0.91m_{\mathrm{TO}}^2+112.07m_{\mathrm{TO}}+1526.01\quad[\mathrm{USD}],\label{eq:drone_price}\\
C_{\mathrm{acq}}=p_{\mathrm{dr}}\sum_jN_j.\label{eq:acquisition}
\end{gather}
% Fig.~\ref{fig:drone_price_fit} shows the data and fitted relationship, with mass expressed in kilograms. The regression is used as a comparative estimate of initial investment, with interpretation constrained by the difference between maximum and design take-off mass and by product-specific differences in structure, certification, and sensors.
% \begin{figure}[!t]
% \centering
% \includegraphics[width=0.85\linewidth]{image_ver3/Drone_price_mass_KO4_compact.pdf}
% \caption{Maximum take-off mass--price relationship and quadratic fit for commercial drones.}
% \label{fig:drone_price_fit}
% \end{figure}
Vehicle price and fleet size are considered together because a lower unit price can be offset by a larger fleet.

%========================================================================================
\section{Seoul Case Study and Results}\label{sec:results}

The framework is applied to Seoul, Republic of Korea, to evaluate how the number of delivery hubs affects vehicle design, operations, and cost. Seoul is a densely populated city with approximately 9.3 million residents within an area of 605 $\mathrm{km}^2$, divided into 25 administrative districts. Comparisons of vehicle designs across different hub configurations and sensitivity analyses are conducted to examine the effects of design and operating assumptions.

\subsection{Delivery Data and Experimental Settings}\label{sec:case_setup}

The case study uses district-level ground delivery records provided by CJ Logistics for June 2022. The dataset contains 4,798,710 deliveries across 625 directed OD pairs among Seoul's 25 districts. The baseline planned OD demand is $K=47,674$ deliveries, obtained by applying the demand scaling factor $\beta=0.01$ in \eqref{eq:demand}.

\begin{figure}[!t]
\centering
\includegraphics[width=\linewidth]{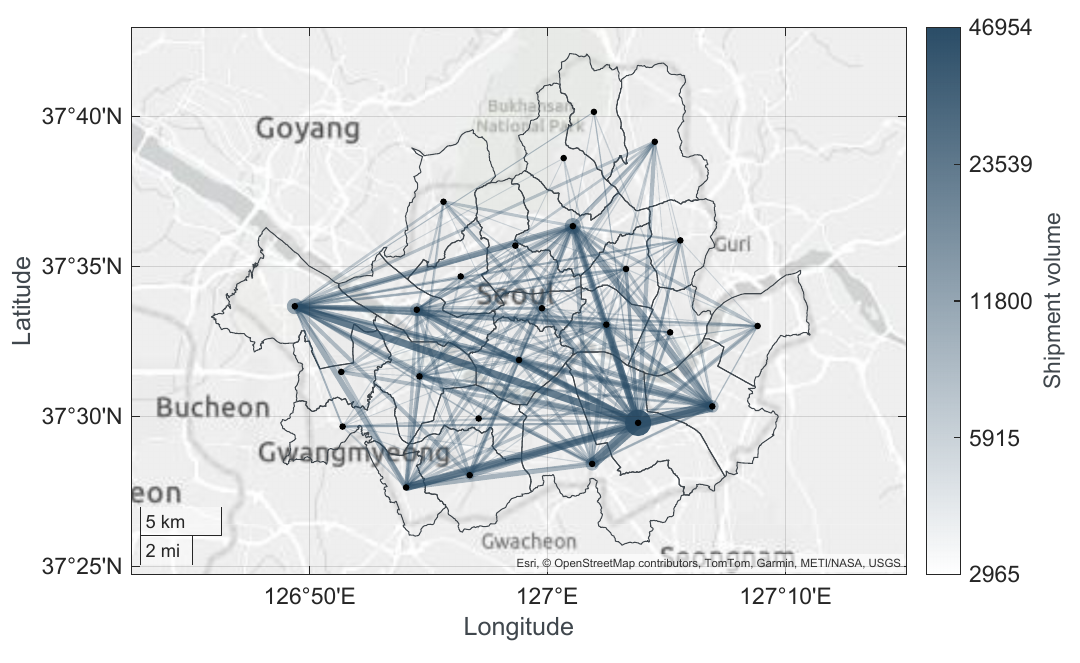}
\caption{Spatial distribution of ground OD demand in Seoul.}
\label{fig:od_spatial}
\end{figure}

Each district's geographic centroid represents both a demand node and a candidate hub location. Inter-node distances are great-circle distances between centroids. Delivery distance within a node is approximated by the distance from the district centroid to its farthest administrative-boundary point and assigned to the corresponding self-loop. Fig.~\ref{fig:od_spatial} shows the spatial distribution of original ground delivery OD demand: node size represents intra-district deliveries, and links show the largest 40\% of inter-district flows.

\begin{figure}[!t]
\centering
\includegraphics[width=0.8\linewidth]{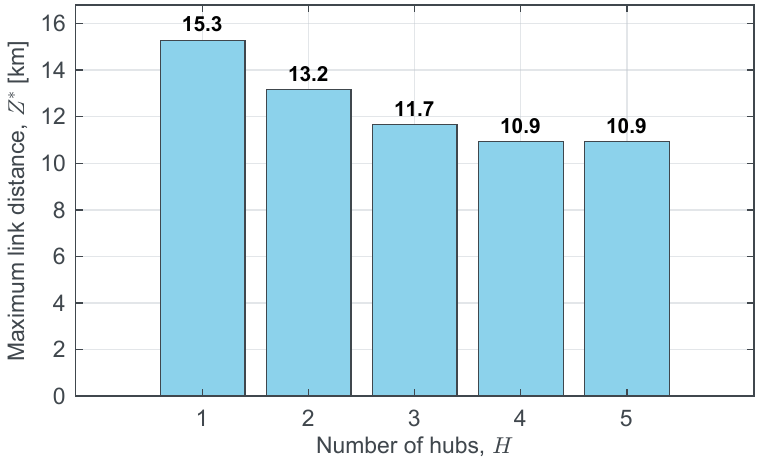}
\caption{Optimal maximum link distance $Z^\ast$ versus hub count.}
\label{fig:hub_count_screening}
\end{figure}

% LHT - important addition
Fig.~\ref{fig:hub_count_screening} shows the optimal maximum link distance $Z^\ast$ for different numbers of hubs using the same 25 candidate sites. The distance decreases from approximately 15.3 km with one hub to 10.9 km with either four or five hubs. The reduction becomes limited at higher hub counts because inter-hub distance also constrains the network. In this study, deliveries between different service zones are required to reach their destinations through at most one hub-to-hub transfer. This requirement prevents the hubs from being placed too far apart and causes the hubs to remain relatively close together, as shown in Fig.~\ref{fig:hubs}. The one- to four-hub scenarios are therefore examined in detail to assess how the reduction in required range, $R_{\mathrm{req}}=2Z^\ast$, affects vehicle design, operations, and cost.

Table~\ref{tab:settings} summarizes the baseline and analysis ranges. Individual package masses are unavailable; a baseline payload of 2 kg is selected with reference to Amazon Prime Air's 5 lb (approximately 2.27 kg) package limit\cite{amazon_prime_air_delivery_updates_2023}. A 10 min hover reserve provides a fixed design energy allowance for flight phases other than cruise. The cruise altitude is assumed to be 300 m, with the corresponding standard-atmosphere conditions used for performance evaluation in every scenario. The battery assumptions are six cells, 22.2 V, and 180 Wh/kg. Monthly operations assume 30 operating days, excluding waiting times for charging, swapping, and maintenance. \rev{Both effective vehicle operating time $h_{\mathrm{day}}$ and staffed hub operating time $h_{\mathrm{hub}}$ are set to 18 h/day at baseline. The hourly rate for all hub staff is set to 12,384 KRW, based on the 2026 Korean minimum wage of 10,320 KRW/h plus a 20\% allowance for paid weekly rest\cite{korea_minimum_wage_2026,moel_paid_weekly_rest}.} Planned OD demand, operating assumptions, and cost assumptions are held fixed across the baseline scenarios, while each scenario's $R_{\mathrm{req}}$ is applied to both route generation and vehicle sizing.
\input{tables_ver3/settings_table.tex}

The hub-area coefficients $a_1=18.5$ and $a_2=80$ m$^2$ are estimated with reference to facility plans published by Wing Aviation LLC, an Alphabet subsidiary operating commercial drone delivery services, for its proposed facilities in Wylie and Mesquite, Texas\cite{wing_wylie_site_plan,mesquite_wing_site_plan_2025}. The estimate for $a_1$ is normalized using reference vehicle dimensions, whereas $a_2$ represents shared fixed space.

A ground-task workload of four person-minutes per flight ($b_1$) is assumed for cargo handling, battery exchange, and preflight checks at the departure hub. This time counts toward staffing but adds no vehicle downtime in \eqref{eq:fleet}. Each hub has one additional staff member ($b_2=1$) for management and operating support.

Rent is estimated using Seoul commercial real-estate information \cite{cbre_seoul_figures_q3_2025}. The same unit rental, utility, and maintenance costs are applied to all hubs. Battery charging electricity is priced at 0.121 USD/kWh, based on the summer mid-load rate of KEPCO's 2025 industrial electricity tariff \cite{kang2025kepco}. This rate excludes basic charges, adjustments, and taxes.

Manufacturer information for selected commercial UAV motors reports mean time between failures or service life of approximately 1,000 h\cite{tmotor_u8lite_kv100,tmotor_u13ii_kv130}. A replacement life of $L_m=800$ h is applied to all designs as a conservative assumption reflecting operating-environment uncertainty; its cost effect is evaluated in the sensitivity analysis.

\subsection{Baseline Network, Vehicle, and Operating Results}\label{sec:baseline}

\begin{figure}[!t]
\centering
\subfloat[1 hub]{\includegraphics[width=.45\linewidth]{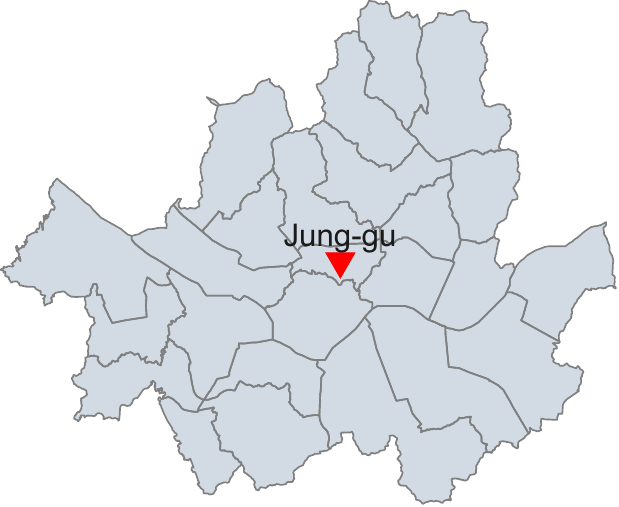}}\hfill
\subfloat[2 hubs]{\includegraphics[width=.45\linewidth]{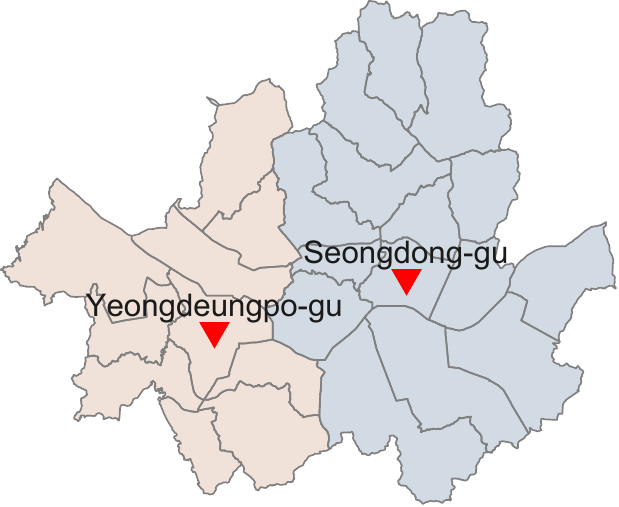}}\\[.6ex]
\subfloat[3 hubs]{\includegraphics[width=.45\linewidth]{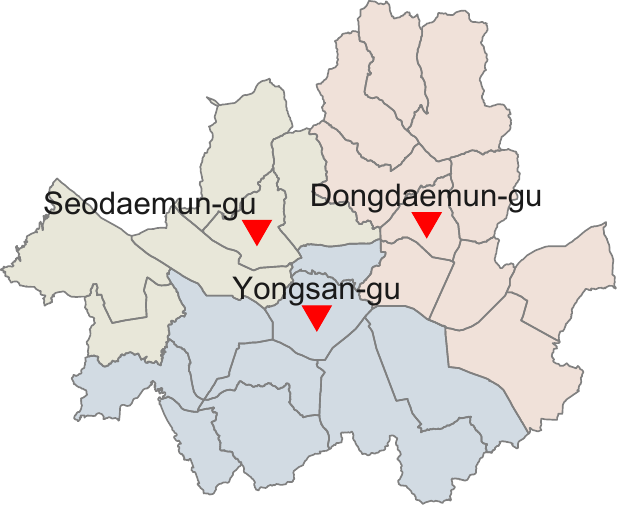}}\hfill
\subfloat[4 hubs]{\includegraphics[width=.45\linewidth]{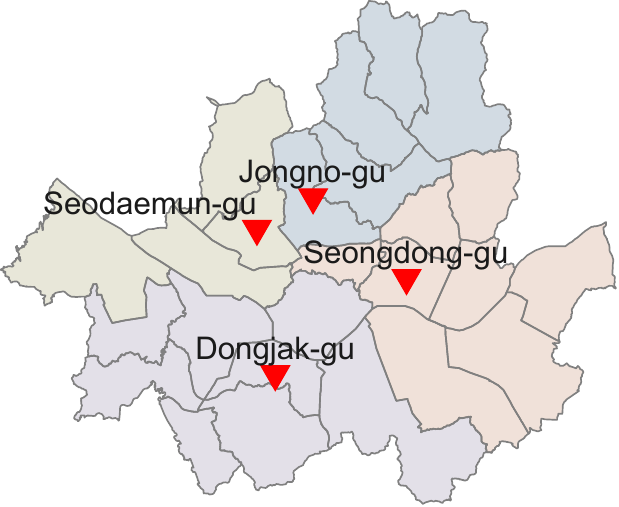}}
\caption{Hub locations and service zones by hub-count scenario (colors: zones; red inverted triangles: hubs).}
\label{fig:hubs}
\end{figure}

The baseline analysis examines whether shorter required ranges from additional hubs translate into lighter vehicles and lower system operating cost. Fig.~\ref{fig:hubs} shows the hub locations and service zones. Increasing hub count from one to four reduces required range from approximately 30.55 to 21.86 km. At the same time, routing inter-zone deliveries through hubs affects flight frequency and the fleet size required at each hub.

\input{tables_ver3/drone_table.tex}

All baseline scenarios select the T-Motor MN5008 KV170 motor with the APC $18\times5.5$MR propeller (Table~\ref{tab:drone}). Nevertheless, battery energy decreases from approximately 721 to 394 Wh and take-off mass from 8.12 to 6.12 kg between the one- and four-hub scenarios, reductions of approximately 45.4\% and 24.7\%, respectively. Even with the same propulsion system and wheelbase, battery sizing and associated mounting and accessory masses produce distinct vehicle designs. The listed MN5008 KV170 unit price is 89.99 USD, used for motor replacement in every baseline scenario.

\begin{figure}[!t]
\centering\includegraphics[width=0.9\columnwidth]{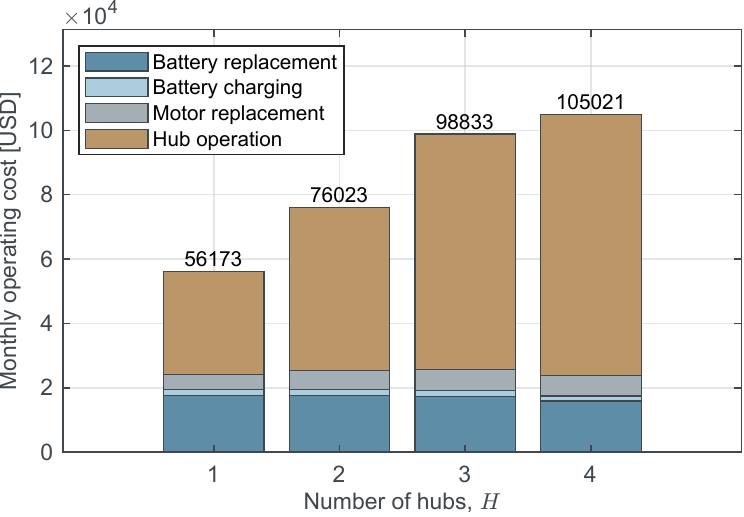}
\caption{Monthly operating cost breakdown by hub-count scenario.}
\label{fig:baselinecost}
\end{figure}

Table~\ref{tab:operation} summarizes the operational requirements and costs, and Fig.~\ref{fig:baselinecost} and Table~\ref{tab:costparts} show the operating cost breakdown. Among vehicle-related operating expenses, battery replacement is the largest cost component, approximately ten times the cost of charging electricity.

\input{tables_ver3/operation_table.tex}

\input{tables_ver3/cost_table.tex}

% LHT - revised with the energy emphasis
Although planned OD demand is unchanged, the total number of flights increases from 28,456 to 52,587, and total flight distance increases by 21.5\%, from 828,931 to 1,007,080 km, between the one- and four-hub scenarios. Nevertheless, total flight energy consumption decreases by 13.1\%, from 15,215 to 13,227 kWh, because the lighter vehicles consume less energy per unit distance. This demonstrates that minimizing total flight distance does not necessarily minimize energy consumption when vehicle designs differ. Despite the reduction in energy consumption, lighter vehicles do not necessarily translate into lower system operating cost. Fleet size increases from 20 to 28, while monthly operating cost increases from approximately \rev{USD 56,173 to USD 105,021}, and operating cost per delivery from \rev{USD 1.18 to USD 2.20}. Increased hub rental and labor costs outweigh the reductions in charging and battery replacement costs. Accordingly, the one-hub configuration has the lowest operating cost among the four baseline configurations.

\rev{Comparing the one- and four-hub scenarios, cost per flight increases slightly from approximately USD 1.97 to USD 2.00, while operating cost per delivery increases more sharply because more flights are required per delivery. Operating cost per delivery therefore accounts for both flight cost and flight requirements.} Initial fleet acquisition cost also increases from approximately USD 49,933 to USD 62,885 because the larger required fleet offsets the lower estimated unit vehicle price.

% LHT need to be careful here
\subsection{Cross-Comparison of Vehicle Designs and Hub Configurations}\label{sec:fixed}

The baseline analysis in Section~\ref{sec:baseline} uses a vehicle specifically designed for each hub configuration. However, a vehicle designed for fewer hubs has a longer flight range and may also be used in a network with more hubs. Although such a vehicle is generally heavier, its longer range may allow more efficient flight routes and reduce the number of required flights. This analysis examines whether these operational benefits can offset the additional costs associated with a heavier vehicle.

A vehicle designed for an $s$-hub network and operated in an $H$-hub network is denoted by $s\to H$. Ten combinations satisfying $1\le s\le H\le4$ are evaluated. For a given $s$, vehicle characteristics remain unchanged, while for a given $H$, hub locations and planned OD demand are fixed. The same operating and cost assumptions are applied to all combinations.

For each combination, the vehicle's achieved range is used as the distance limit to regenerate candidate routes, transform segment demand, and optimize flight frequencies. Cases with $s=H$ correspond to the baseline results.

The updated flight plan and the vehicle's cruise speed are used to recalculate the required fleet size at each hub. \rev{Fleet size and vehicle wheelbase determine hub space, while hub-level flight frequency determines concurrent staffing in \eqref{eq:hubcostsize}. Rent and labor costs are recalculated accordingly.} These costs are then combined with charging and component replacement costs to obtain the total operating cost.

\begin{figure}[!t]
\centering
\subfloat[{Operating cost per delivery}]{\includegraphics[width=.9\linewidth]{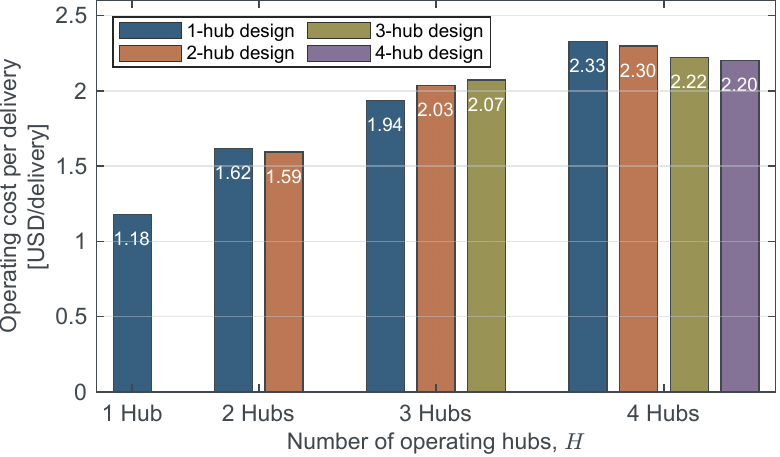}}\par
\subfloat[Total flights]{\includegraphics[width=.9\linewidth]{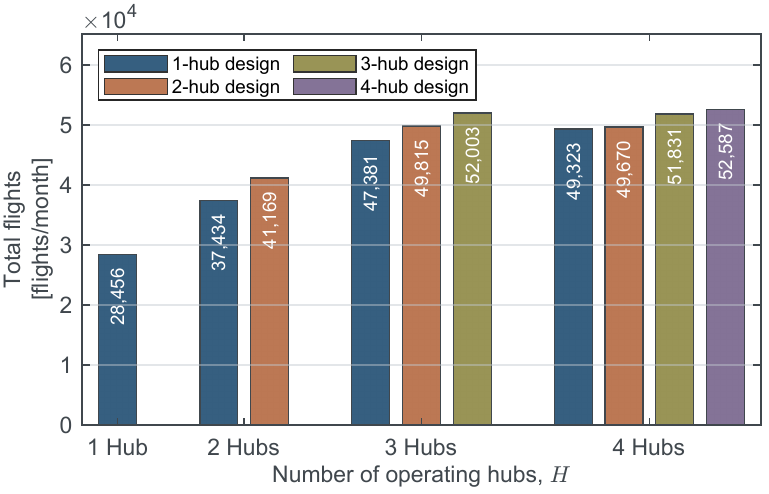}}\par
\subfloat[Total energy consumption]{\includegraphics[width=.9\linewidth]{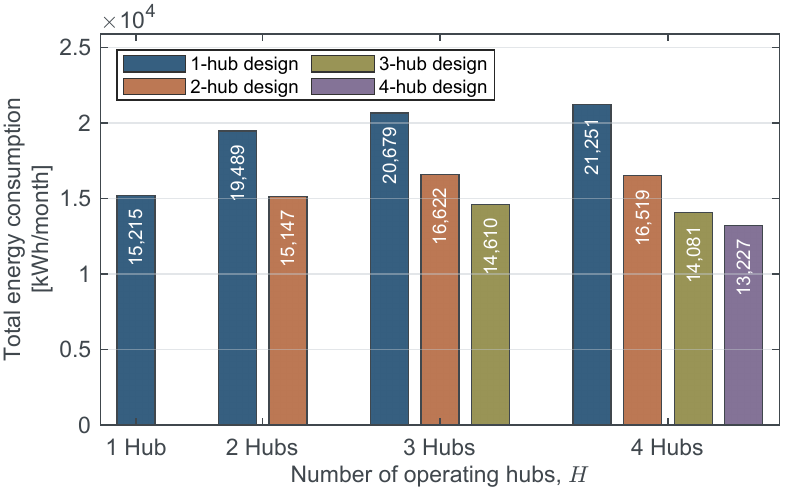}}
\caption{Monthly operating results by vehicle design and operating hub count (colors: vehicle design $s$).}
\label{fig:comparison}
\end{figure}

Fig.~\ref{fig:comparison} and Table~\ref{tab:comparison} compare the resulting vehicle--network combinations. Each group along the figure's horizontal axis represents an operating hub count $H$, with colors identifying the design scenario $s$.
\input{tables_ver3/comparison_main5_table.tex}

\rev{In the two- and four-hub networks, the vehicles designed for the corresponding hub configurations ($2\to2$ and $4\to4$) have the lowest operating costs. In the three-hub network, however, the one-hub vehicle ($1\to3$) achieves approximately 1.937 USD per delivery, 6.59\% lower than the 2.073 USD of the three-hub vehicle ($3\to3$).}

\rev{Across the two- to four-hub networks, the one-hub design requires the fewest flights but consumes the most energy: its longer range permits more routes, while its greater mass increases power demand. In the three-hub network, using this design instead of the three-hub design reduces monthly flights from 52,003 to 47,381 and total concurrent staffing across the hubs from 12 to 9. Although the fleet grows from 27 to 28 drones of the same wheelbase, the labor savings outweigh the increases in rent and energy-related costs.}

% LHT important summary
\rev{Overall, minimum-mass sizing can reduce energy consumption, but does not necessarily minimize operating cost. A longer-range vehicle can reduce flight frequency and associated staffing costs. These findings support evaluating vehicle design together with network requirements and ground operations.}

\subsection{Propulsion Selection Under Different Mission Requirements}\label{sec:designgrid}

The Seoul baseline selects the same MN5008 KV170 motor and $18\times5.5$MR propeller for all four hub-count scenarios at a 2 kg payload. To examine the range of conditions over which this identical propulsion selection persists, designs are compared at 30 combinations of payloads from 0.5 to 2.5 kg and required ranges from 10 to 35 km. 

\begin{figure}[!t]
\centering
\includegraphics[width=\linewidth]{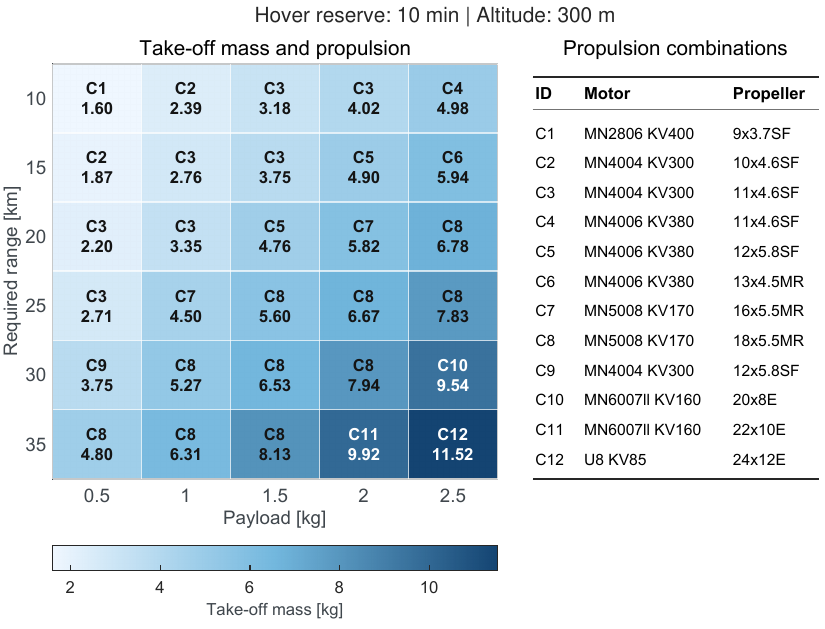}
\caption{Take-off mass and propulsion configuration by range and payload.}
\label{fig:design_grid}
\end{figure}

Fig.~\ref{fig:design_grid} combines take-off mass and propulsion selection. Each cell shows a configuration ID above the mass in kilograms; background color also represents mass. IDs C1--C12 identify the motor--propeller combinations listed on the right, with identical IDs indicating identical propulsion systems.

Both persistent and changing propulsion selections occur across the evaluated conditions. At a 2 kg payload, increasing required range from 20 to 25 km changes the selected propeller on the MN5008 KV170 motor from $16\times5.5$MR (C7) to $18\times5.5$MR (C8). At 25 and 30 km, C8 remains selected, as in the baseline, while battery sizing increases take-off mass from approximately 6.67 to 7.94 kg.

Although the candidate database includes conventional propellers, the optimization generally selects multirotor (MR) or other low-pitch propellers. This is consistent with the low-speed, high-thrust requirements of the delivery mission and supports the physical reasonableness of the resulting designs.

\subsection{Sensitivity of Operating Costs to Demand and Operating Assumptions}\label{sec:sensitivity}

The baseline analysis estimates operating costs of \rev{1.18--2.20 USD per delivery} under the assumed demand and operating conditions. However, several cost parameters are based on estimates that may differ from actual commercial operations. Sensitivity analysis is therefore conducted to examine how these assumptions influence the estimated costs and to assess the robustness of the baseline results.

The analysis examines the effects of planned OD demand, \rev{daily operating hours}, component lifetimes, and hub-area and staffing assumptions. Operating cost per delivery is used to evaluate changes in cost levels and the relative economic performance of different hub configurations.

Each parameter is varied individually, with the vehicle designs and other assumptions held fixed. When planned OD demand changes, the original demand matrix is scaled uniformly, preserving its relative distribution apart from rounding. Segment demand and flight frequencies are then recalculated. For the remaining parameters, the network and flight plan are unchanged, while the affected operating requirements and costs are reevaluated. \rev{The operating-hours analysis varies effective vehicle operating time and staffed hub operating time together ($h_{\mathrm{day}}=h_{\mathrm{hub}}$).}

\begin{figure}[!t]
\centering
\subfloat[{Demand scaling factor}]{\includegraphics[width=.48\linewidth]{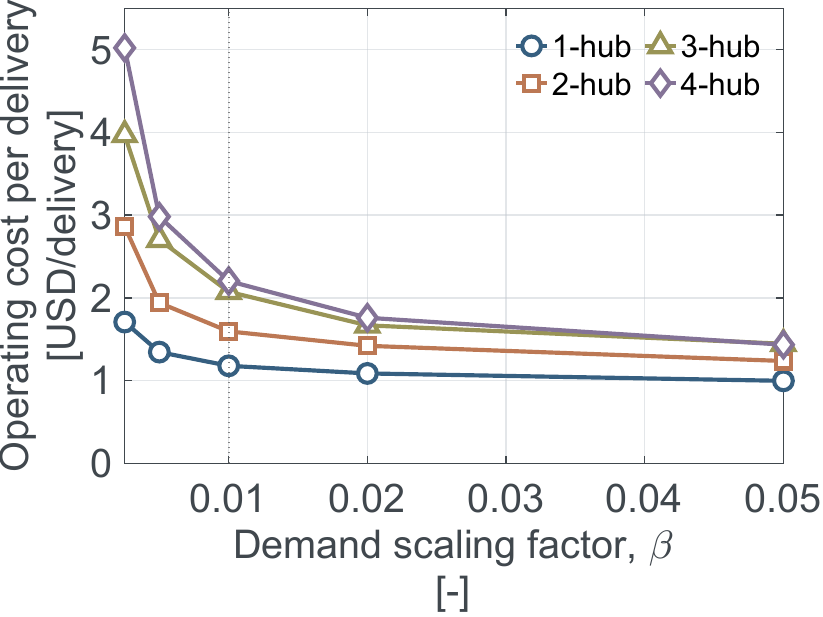}}\hfill
\subfloat[Daily operating hours]{\includegraphics[width=.48\linewidth]{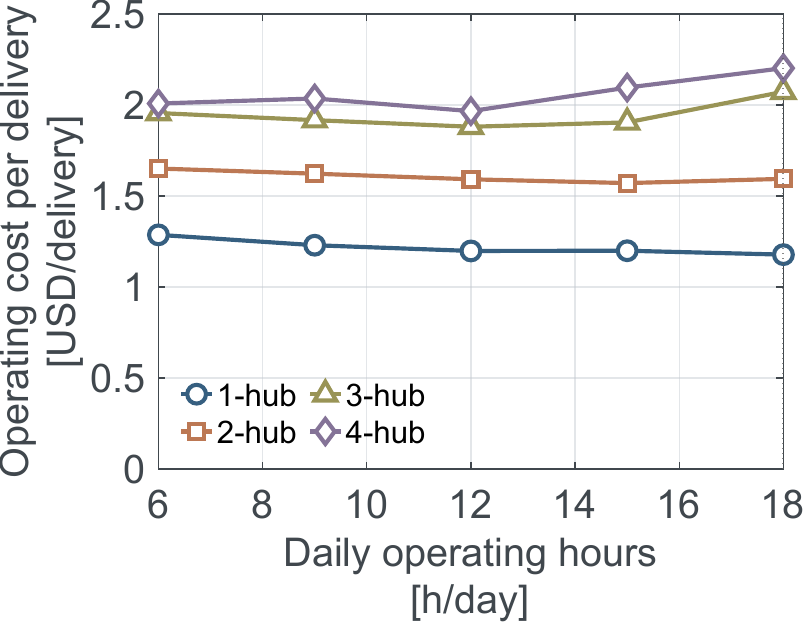}}\\[.6ex]
\subfloat[Battery cycle life]{\includegraphics[width=.48\linewidth]{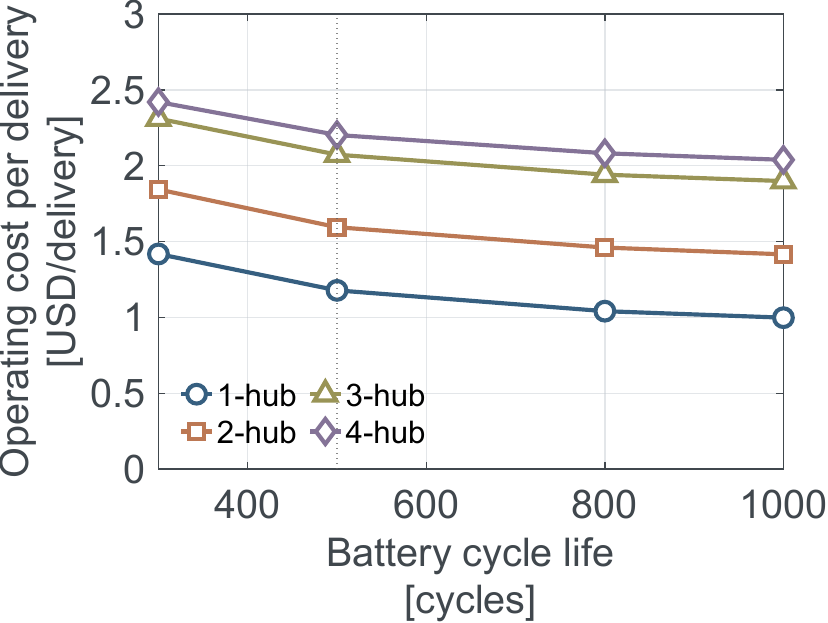}}\hfill
\subfloat[Motor service life]{\includegraphics[width=.48\linewidth]{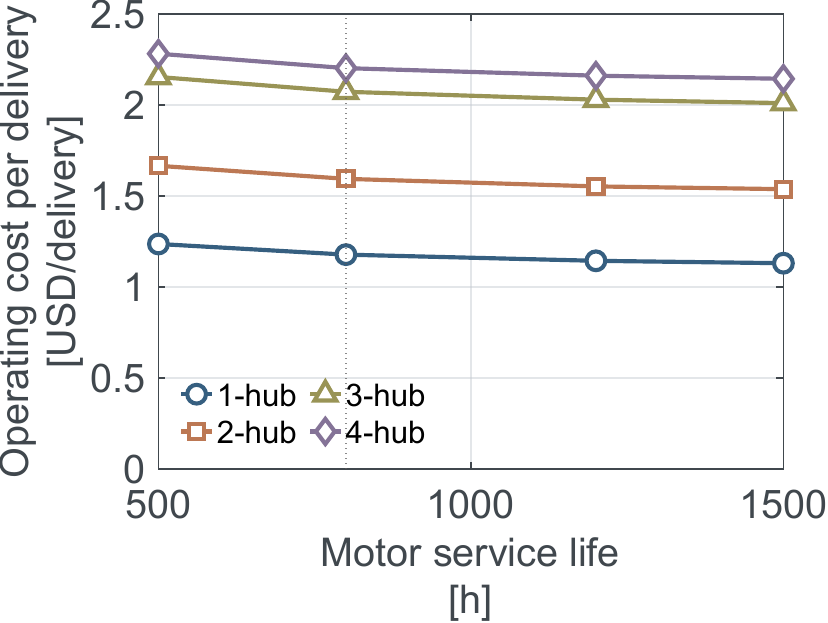}}
\caption{Sensitivity of {operating cost per delivery} to {the demand scaling factor}, daily operating hours, and component lifetimes.}
\label{fig:sensoperation}
\end{figure}

\begin{figure}[!t]
\centering
\subfloat[Area coefficient $a_1$]{\includegraphics[width=.48\linewidth]{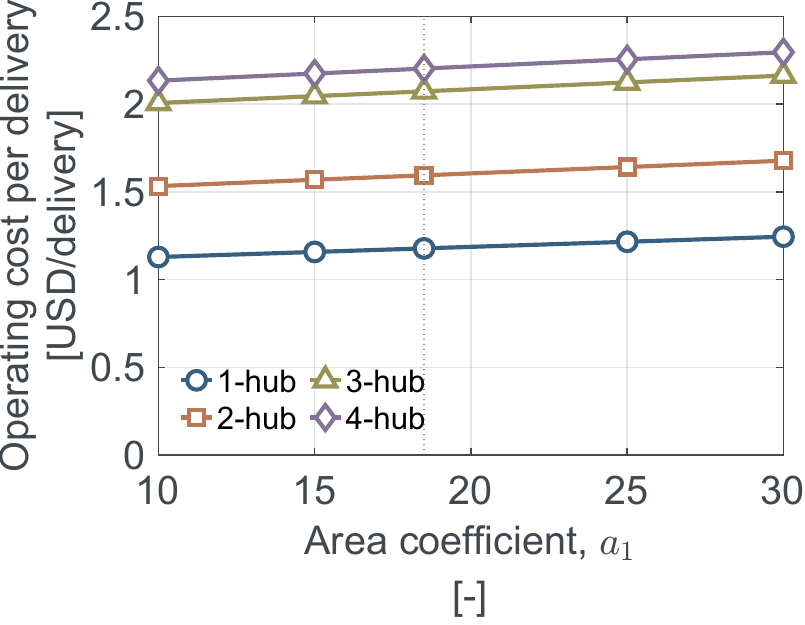}}\hfill
\subfloat[Shared fixed hub area $a_2$]{\includegraphics[width=.48\linewidth]{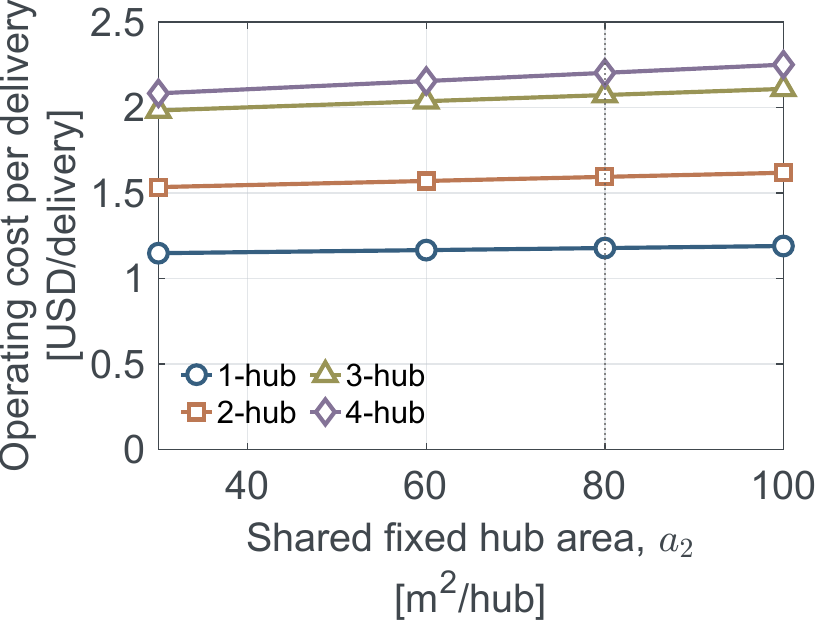}}\\[.6ex]
\subfloat[Ground-task labor $b_1$]{\includegraphics[width=.48\linewidth]{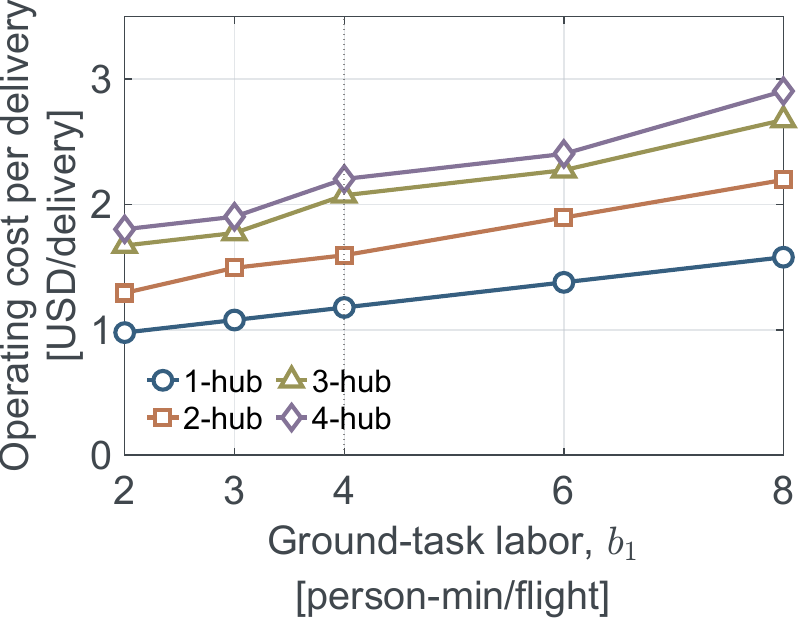}}\hfill
\subfloat[Base hub staffing $b_2$]{\includegraphics[width=.48\linewidth]{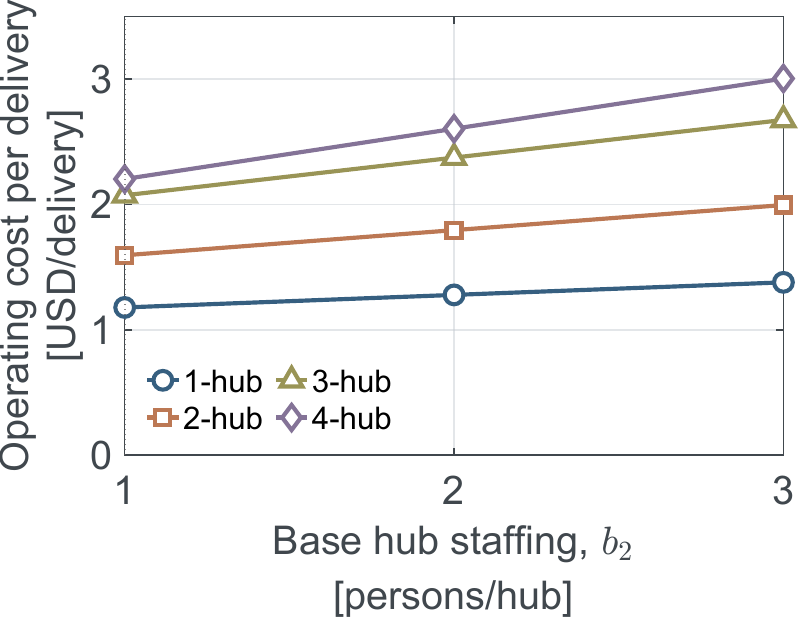}}
\caption{Sensitivity of {operating cost per delivery} to {hub-area and staffing coefficients}.}
\label{fig:senshub}
\end{figure}

Figs.~\ref{fig:sensoperation} and \ref{fig:senshub} show the resulting operating cost per delivery. Markers denote evaluated points, dashed reference lines indicate baseline inputs, and connecting lines are visual guides. Because the input ranges differ, interpretation focuses on changes in operating cost per delivery within each evaluated range rather than ranking the importance of the inputs.


% This needs to be rewritten LHT
The analysis first examines the effect of monthly delivery demand on operating cost per delivery. Demand is increased by applying a common scaling factor $\beta$ to the original OD matrix, approximately preserving the relative distribution among OD pairs. As monthly demand increases from 11,683 to 239,637 deliveries, operating cost per delivery decreases across all four hub configurations (Fig.~\ref{fig:sensoperation}(a)). Operating cost per delivery decreases rapidly at lower demand levels and gradually levels off as demand increases. Although energy consumption and fleet-dependent costs increase with demand, fixed hub costs are distributed over more deliveries, resulting in economies of scale. For example, in the one-hub scenario, monthly operating cost increases from approximately \rev{19,948 to 238,841 USD}, while operating cost per delivery decreases from \rev{1.71 to 1.00 USD}.

Monthly delivery demand also affects the cost rankings of hub configurations. At baseline, the three-hub configuration is less costly than the four-hub configuration. However, at the highest evaluated demand of 239,637 deliveries per month, their operating costs are \rev{1.45 and 1.43 USD per delivery}, respectively, reversing their ranking. The one-hub configuration remains the least costly across all evaluated demand levels.

Fig.~\ref{fig:sensoperation}(b)--(d) shows the effects of operating hours and component lifetimes on operating cost per delivery. The vehicle designs and flight plans remain unchanged in these comparisons. \rev{Longer daily operation reduces fleet size and rent, but extends the duration of base staffing and can increase paid person-hours after integer allocation. Total cost is therefore nonmonotonic over the evaluated hours. The lowest evaluated costs occur at 18, 15, 12, and 12 h/day for the one- through four-hub configurations, respectively; these are comparisons among sampled hours, not optimized schedules.} Longer battery and motor lives reduce replacement costs.

Fig.~\ref{fig:senshub} shows the effects of hub-area and staffing coefficients on operating cost per delivery. \rev{Area coefficients affect rent, while ground-task labor and base staffing determine paid person-hours. Increasing $b_2$ from one to three persons per hub raises four-hub cost from approximately 2.20 to 3.00 USD per delivery. The effects of $a_1$ depend on fleet size, whereas those of $b_1$ depend on flight frequency; shared fixed area $a_2$ and base staffing $b_2$ have greater effects with more hubs.} The one-hub configuration remains the least costly throughout these comparisons, indicating that the baseline cost ranking is robust to the evaluated hub-area and staffing assumptions.

\rev{Integer fleet allocation, staffing, and component replacement create stepwise changes in operating costs. The sensitivity analysis distinguishes changes in cost levels from changes in scenario rankings: the one-hub configuration remains the least costly across the evaluated ranges, although the three- and four-hub rankings reverse at the highest demand. The framework can therefore be used to compare the cost consequences of operating assumptions before detailed scheduling and facility design.}

%\subsection{Framework Applications and Scope}\label{sec:discussion}
%\rev{The framework helps identify cost-efficient hub--vehicle combinations among those evaluated and explain the sources of their cost differences. Other regions can be examined with local spatial and OD data, candidate hubs, and unit prices, provided that the delivery-system assumptions are appropriate. Local facility and staffing evidence is particularly important for selecting hub-area and staffing coefficients; transferability beyond Seoul remains to be established through additional case studies.}

%\rev{The results are optimistic planning-level cost estimates under the stated assumptions, not mathematical lower bounds or predictions of commercial expenditure. Monthly aggregate planning simplifies parcel-level delivery and loading constraints, with fleet requirements estimated from monthly flight time using the assumed 18 h of available daily operating time per vehicle. Operational waiting and broader ownership costs remain outside the monthly cost scope. Because sequential optimization is distinct from jointly minimizing total operating cost over network, routing, and vehicle decisions, the cost rankings apply to the evaluated vehicle--network combinations. Future work should address field validation, time-dependent demand, detailed operational constraints, and full ownership costs.}

% VI.F contents are inserted in conclusions

% LHT energy 
\section{Conclusions}\label{sec:conclusions}

A sequential framework was developed to connect delivery network design, flight-route optimization, mission-specific vehicle sizing, and operating cost assessment. The framework derives vehicle range requirements from hub locations and uses the resulting vehicle characteristics and flight plans to estimate fleet requirements and operating costs. Its application to Seoul demonstrates the importance of considering the relationships between network configuration, vehicle design, and delivery operations.

Under the Seoul baseline conditions, increasing the number of hubs from one to four reduced the required flight range from 30.55 to 21.86 km and vehicle take-off mass from 8.12 to 6.12 kg. Although the number of required flights increased from 28,456 to 52,587 per month and total flight distance increased by 21.5\%, total flight energy consumption decreased by 13.1\%, from 15.2 to 13.2 MWh. However, operating cost per delivery increased from \rev{USD 1.18 to USD 2.20}. Battery replacement was the largest vehicle-related operating expense, while hub rent and labor costs dominated the differences between network configurations. The one-hub configuration was therefore the least costly among the evaluated baseline alternatives, as increased hub operating expenses outweighed the savings associated with lighter vehicles and lower energy consumption. These results demonstrate that reductions in vehicle mass and energy consumption do not necessarily lead to lower operating costs, highlighting the importance of evaluating both energy consumption and operating cost in delivery system design.

Comparisons across vehicle designs further demonstrated that minimum-mass sizing and minimum operating cost need not coincide. Vehicles optimized for the required range achieved the lowest operating costs in the two- and four-hub networks. In the three-hub network, however, the vehicle designed for the one-hub network reduced operating cost by 6.59\% compared with the vehicle optimized for the three-hub network. Its longer range reduced flight frequency and staffing sufficiently to outweigh higher energy-related costs. These results reinforce the importance of evaluating vehicle design together with network requirements and ground operations.

\rev{The sensitivity analysis showed that higher delivery demand reduced operating cost per delivery through economies of scale. Longer daily operation involved a trade-off between lower fleet requirements and longer staffing duration, rather than a uniform cost reduction. Although cost levels changed with the operating and staffing assumptions, the one-hub configuration remained the least costly across all evaluated ranges.}

The framework assumes that planned OD demand is known a priori and that drones fly direct paths between nodes. In practice, altitude limits, airspace restrictions, obstacle avoidance, and ground-risk constraints in dense urban areas may lengthen actual flight paths. \rev{The staffing estimate uses average flight rates and aggregate person-hours, without explicit peak-demand coverage or shift scheduling. The labor rate includes the assumed weekly-rest allowance but excludes other wage premiums and employer on-costs.} Actual commercial operating costs are likely to be higher, because the cost model also excludes expenses such as vehicle depreciation, insurance, and traffic management fees. The framework is thus more suitable for comparative planning than for predicting actual delivery costs.  It can be applied to other regions using local geographic and delivery demand data, together with appropriate operating and cost assumptions.

The framework can also be used to evaluate the potential effects of ground-operation automation and battery characteristics, particularly the trade-off between battery specific energy and cycle life, on vehicle design and operating costs.

\section*{Acknowledgments}
This work was supported by a National Research Foundation of Korea grant funded by the Ministry of Science and ICT (RS-2022-NR070854) and by the Ministry of Land, Infrastructure and Transport through its Data Driven Air Traffic Management program (RS-2021-KA163373).

The authors used OpenAI ChatGPT to edit the manuscript for grammar, clarity, and organization, and OpenAI Codex for Korean-to-English translation and to redraw and format their original illustrations (Figs. 1–7). The methods, analyses, and results are the authors’ own.

% This work was supported by the National Research Foundation of Korea (NRF) grant funded by the Korea government [Ministry of Science and ICT (MSIT)] under Grant RS-2022-NR070854, and by the Data Driven Air Traffic Management program funded by the Ministry of Land, Infrastructure and Transport of the Korean government under Grant RS-2021-KA163373.

% % Before submission, reconcile the disclosure with all retained AI-assisted content and code.
% OpenAI ChatGPT was used to assist with English language editing, grammar checking, and improving the clarity and organization of the manuscript. OpenAI Codex was used to assist with Korean-to-English translation and with redrawing and formatting Figs.~\ref{fig:framework}--\ref{fig:sizing} based on original author generated figures. All research methods, analyses, and results were developed by the authors. 

\bibliographystyle{IEEEtran}
% Highlight only the newly added reference; retain all existing revision marks.
% \begingroup
% \let\unmarkedbibitem\bibitem
% \RenewDocumentCommand{\bibitem}{o m}{%
%   \normalcolor
%   \def\currentbibkey{#2}\def\addedbibkey{bao2026}%
%   \ifshowrevisions\ifx\currentbibkey\addedbibkey\color{red}\fi\fi
%   \IfNoValueTF{#1}{\unmarkedbibitem{#2}}{\unmarkedbibitem[#1]{#2}}%
% }
\bibliography{Main_KO4_refs}
% \endgroup
\end{document}

%% file: tables_ver3/stages_table.tex
% Shared numerical table for Main_KO4.tex and Main_EN_TITS.tex.
\begin{table}[!t]
\centering\caption{Stage-specific decisions and evaluation criteria.}
\label{tab:stages}\footnotesize\renewcommand{\arraystretch}{1.22}
\setlength{\tabcolsep}{3pt}
\begin{tabularx}{\linewidth}{@{}>{\raggedright\arraybackslash}p{1.35cm}YY@{}}
\toprule
Stage & Decision or evaluation criterion & Key conditions\\
\midrule
Hub location & Minimize maximum link distance; break ties by mean hub-assignment distance over all demand nodes & Fixed hub count, candidate sites, and single-hub assignment\\
Route planning & Minimize total flights over generated candidate routes & Segment demand coverage, hub return, and range limit\\
Vehicle sizing & Minimize take-off mass subject to mission requirements & Required range, payload, {hover reserve time}, and component candidates\\
Operating cost assessment & Calculate monthly operating cost and {operating cost per delivery} & Selected vehicle and flight plan; {fixed operating assumptions and unit prices}\\
\bottomrule
\end{tabularx}
\end{table}

%% file: tables_ver3/mass_constants_table.tex
% Shared numerical table for Main_KO4.tex and Main_EN_TITS.tex.
\begin{table}[!t]
\centering\footnotesize
\caption{Design constants and fixed assumptions for vehicle sizing.}
\label{tab:mass_constants}
\begin{tabularx}{\columnwidth}{cYr}
\toprule
\multicolumn{1}{l}{Symbol} & Definition and application & \multicolumn{1}{l}{Value}\\\midrule
$\phi$ & Angle between adjacent rotor arms [deg] & 90\\
$g_{\mathrm{disc}}$ & Rotor clearance / propeller diameter & 0.075\\
$r_{\mathrm{mf}}$ & Main-frame diameter / wheelbase & 0.35\\
$r_{\mathrm{al}}$ & Rotor-arm length / wheelbase & 0.45\\
$r_{\mathrm{skid}}$ & Skid outer diameter / rotor-arm outer diameter & 0.7\\
$k_{\mathrm{wr}}$ & Mass-reduction factor for frame and mount plates & 0.8\\
$\rho_{\mathrm{cf}}$ & Structural carbon-material density [kg/m$^3$] & 1500\\
$t_{\mathrm{plate}}$ & Plate thickness: main frame, motor mounts, and battery mount [mm] & 2\\
$t_{\mathrm{bar}}$ & Tube-wall thickness: rotor arms and landing skids [mm] & 1\\
$N_{\mathrm{mp}}$ & Number of main-frame plates & 2\\
$N_{b,\mathrm{pack}}$ & Number of battery packs for mount sizing & 1\\
$m_{\mathrm{av}}$ & Avionics mass [kg] & 0.18\\
$k_{\mathrm{wire}}$ & Wiring mass / propulsion-system mass & 0.1\\
$k_{\mathrm{subs}}$ & Accessory mass / structural mass & 0.1\\\bottomrule
\end{tabularx}
\par\smallskip
\begin{minipage}{\columnwidth}\scriptsize
\leavevmode{Note: Electronic speed controller (ESC) mass is not accounted for separately.}
\end{minipage}
\end{table}

%% file: tables_ver3/aero_constants_table.tex
% Shared numerical table for Main_KO4.tex and Main_EN_TITS.tex.
\begin{table}[!t]
\centering\footnotesize
\caption{Aerodynamic and geometric inputs for the forward-flight model.}
\label{tab:aero_constants}
\begin{tabularx}{\columnwidth}{cYr}
\toprule
\multicolumn{1}{l}{Symbol} & Description & \multicolumn{1}{l}{Value}\\\midrule
$C_L$ & Downward vertical force coefficient & 0.193\\
$C_D$ & Vehicle drag coefficient & 0.425\\
$m_{\mathrm{ref}}$ & Reference vehicle mass [kg] & 13.5\\
$S_{\mathrm{top,ref}}$ & Top-view projected area [m$^2$] & $0.585^2$\\
$S_{\mathrm{front,ref}}$ & Front-view projected area [m$^2$] & $0.585\times0.4$\\
$S_{\mathrm{front,box}}$ & Payload frontal area [m$^2$] & $0.3^2$\\\bottomrule
\end{tabularx}
\end{table}

%% file: tables_ver3/commercial_table.tex
% Shared numerical table for Main_KO4.tex and Main_EN_TITS.tex.
\begin{table*}[!t]
\centering\caption{{Commercial drone specifications and corresponding model-based designs.}}
\label{tab:commercial}\footnotesize\renewcommand{\arraystretch}{1.18}
\setlength{\tabcolsep}{4.5pt}
% Content-width shares: 23.5%, 16.5%, 10%, 17%, 7.5%, 12.75%, and 12.75%.
\begin{tabularx}{\textwidth}{@{}
>{\hsize=1.645\hsize\linewidth=\hsize\raggedright\arraybackslash}X
>{\hsize=1.155\hsize\linewidth=\hsize\raggedright\arraybackslash}X
>{\hsize=0.70\hsize\linewidth=\hsize\centering\arraybackslash}X
>{\hsize=1.19\hsize\linewidth=\hsize\centering\arraybackslash}X
>{\hsize=0.525\hsize\linewidth=\hsize\raggedleft\arraybackslash}X
>{\hsize=0.8925\hsize\linewidth=\hsize\raggedleft\arraybackslash}X
>{\hsize=0.8925\hsize\linewidth=\hsize\raggedleft\arraybackslash}X@{}}
\toprule
Aircraft / source & Motor & \multicolumn{1}{l}{{Propeller [in]}} & \multicolumn{1}{l}{{Battery inputs}} & \multicolumn{1}{l}{{Mass [kg]}} & \multicolumn{1}{l}{{Wheelbase [mm]}} & \multicolumn{1}{l@{}}{{Endurance [min]}}\\\midrule
Matrice 100 specifications\cite{djiMatrice100Specs} & DJI 3510 KV350 & $13\times4.5$ & \multirow{2}{*}{{\makecell[c]{600 g; 6S\\4,500 mAh; 99.9 Wh}}} & 2.355 & 650 & 22\\
\leavevmode{Model-based design} & MN3510 KV360 & $13\times4.5$MR & & 1.825 & 502 & 32\\\midrule
Inspire 1 V2 specifications\cite{djiInspire1Specs} & DJI 3510H KV420 & $13\times4.5$ & \multirow{2}{*}{{\makecell[c]{570 g; 6S\\4,500 mAh; 99.9 Wh}}} & 2.845 & 581 & 18\\
\leavevmode{Model-based design} & MN3520 KV400 & $13\times4.5$MR & & 2.254 & 502 & 20\\\midrule
M690A retailer listing\cite{rcdroneM690AListing} & T-Motor 5005 KV520 & $18\times6$ & \multirow{2}{*}{{\makecell[c]{2,100 g; 4S\\30,000 mAh; 444 Wh}}} & 3.250$^a$ & 699 & $\ge71^b$\\
\leavevmode{Model-based design} & MN5006 KV450 & $18\times5.5$MR & & 3.923 & 695 & 64\\\midrule
Mavic 3T specifications\cite{djiMavic3EnterpriseSpecs} & DJI 2008 KV1400 & $9.4\times5.3$ & \multirow{2}{*}{{\makecell[c]{335.5 g; 4S\\5,000 mAh; 77 Wh}}} & 0.920 & 380 & 38\\
\leavevmode{Model-based design} & AS2306 KV1500 & $9.5\times5$MR & & 0.930 & 367 & 24\\\bottomrule
\end{tabularx}
\par\smallskip
\begin{minipage}{.96\textwidth}\scriptsize
% \leavevmode{Note: Voltage [V] = energy [Wh]/capacity [Ah]; specific energy [Wh/kg] = energy [Wh]/mass [kg].}\\
$^a$ {Estimated zero-payload mass: retailer-listed 4.25 kg minus 1 kg payload; not independently verified.}\\
$^b$ Endurance with a 0.5 kg payload, not zero-payload hover time.
\end{minipage}
\end{table*}

%% file: tables_ver3/settings_table.tex
% KO4 notation copy; numerical values preserved from ko2_assets.
% Generated from audited MATLAB results by Re_ExportPaperKO2.m
\begin{table}[!t]
\centering
\caption{{Baseline inputs} and analysis ranges.}
\label{tab:settings}
\footnotesize
\setlength{\tabcolsep}{2.5pt}
\renewcommand{\arraystretch}{1.18}
\begin{tabularx}{\columnwidth}{@{}>{\raggedright\arraybackslash}p{0.40\columnwidth}>{\centering\arraybackslash}p{0.19\columnwidth}Y@{}}
\toprule
Input & {\raggedright Baseline\par} & Analysis range \\
\midrule
Payload [kg] & 2 & 0.5, 1, 1.5, 2, 2.5\newline (design grid) \\
Hover reserve time [min] & 10 & Fixed \\
\leavevmode{Cruise altitude [m]} & 300 & Fixed \\
\leavevmode{Demand scaling factor $\beta$} & 0.01 & 0.0025, 0.005, 0.01, 0.02, 0.05 \\
Daily operating hours\newline $h_{\mathrm{day}}=h_{\mathrm{hub}}$ [h/day] & 18 & 6, 9, 12, 15, 18 \\
Battery cycle life [cycles] & 500 & 300, 500, 800, 1,000 \\
Motor service life [h] & 800 & 500, 800, 1,200, 1,500 \\
Area coefficient $a_1$ & 18.5 & 10, 15, 18.5, 25, 30 \\
Shared fixed hub area $a_2$\newline [m$^2$/hub] & 80 & 30, 60, 80, 100 \\
Ground-task labor $b_1$\newline [person $\cdot$ min/flight] & 4 & 2, 3, 4, 6, 8 \\
Base hub staffing $b_2$\newline [persons/hub] & 1 & 1, 2, 3 \\
Electricity price {[USD/kWh]} & \leavevmode{0.121} & Fixed \\
Rent {[USD/m$^2$/month]} & \leavevmode{28.57} & Fixed \\
Labor rate $w_L$\newline [USD/(person $\cdot$ h)] & 8.846 & Fixed \\
Hub utilities / maintenance\newline {[USD/hub/month]} & \makecell[c]{{357.1} /\\{357.1}} & Per hub \\
\bottomrule
\end{tabularx}
\end{table}

%% file: tables_ver3/drone_table.tex
% KO4 one-column layout; display values rounded from the audited saved results.
% Keep the shared generated asset unchanged for earlier manuscripts.
\begin{table}[!t]
\centering
\caption{Vehicle design results by hub count $H$.}
\label{tab:drone}
\footnotesize
\setlength{\tabcolsep}{3pt}
\renewcommand{\arraystretch}{1.18}
\begin{tabularx}{\linewidth}{@{}Yrrrr@{}}
\toprule
Parameter & \multicolumn{1}{l}{$H=1$} & \multicolumn{1}{l}{$H=2$} & \multicolumn{1}{l}{$H=3$} & \multicolumn{1}{l@{}}{$H=4$} \\
\midrule
Motor & \multicolumn{4}{c}{MN5008 KV170 (all scenarios)} \\
Propeller & \multicolumn{4}{c}{$18\times5.5$MR (all scenarios)} \\
Required range [km] & 30.55 & 26.33 & 23.33 & 21.86 \\
Achieved range [km] & 30.55 & 26.33 & 23.33 & 21.86 \\
Battery energy [Wh] & 721 & 530 & 434 & 394 \\
Battery mass [kg] & 4.01 & 2.95 & 2.41 & 2.19 \\
Motors + propellers (4 sets) [kg] & 0.77 & 0.77 & 0.77 & 0.77 \\
Airframe [kg] & 0.630 & 0.627 & 0.625 & 0.625 \\
Avionics [kg] & 0.18 & 0.18 & 0.18 & 0.18 \\
Payload [kg] & 2.00 & 2.00 & 2.00 & 2.00 \\
Miscellaneous [kg] & 0.54 & 0.43 & 0.38 & 0.36 \\
Take-off mass $m_{\mathrm{TO}}$ [kg] & 8.12 & 6.95 & 6.36 & 6.12 \\
Wheelbase [m] & 0.70 & 0.70 & 0.70 & 0.70 \\
Cruise speed [m/s] & 21.98 & 21.38 & 20.88 & 20.45 \\
\bottomrule
\end{tabularx}
\par\smallskip
\parbox{\linewidth}{\scriptsize \leavevmode{Note:} Values are rounded; component masses may not sum to the displayed take-off mass.}
\end{table}

%% file: tables_ver3/operation_table.tex
% KO4 one-column layout; selected rows retain values from ko2_assets/operation_table.tex.
\begin{table}[!t]
\centering
\caption{Monthly operating results and initial vehicle acquisition cost by hub count $H$.}
\label{tab:operation}
\footnotesize
\setlength{\tabcolsep}{3pt}
\renewcommand{\arraystretch}{1.18}
\begin{tabularx}{\linewidth}{@{}Yrrrr@{}}
\toprule
Metric & \multicolumn{1}{l}{$H=1$} & \multicolumn{1}{l}{$H=2$} & \multicolumn{1}{l}{$H=3$} & \multicolumn{1}{l@{}}{$H=4$} \\
\midrule
Planned OD demand $K$ [deliveries] & 47,674 & 47,674 & 47,674 & 47,674 \\
Total flights $F$ [flights] & 28,456 & 41,169 & 52,003 & 52,587 \\
Fleet size $\sum_jN_j$ [drones] & 20 & 25 & 27 & 28 \\
Total flight distance $10^{-3}\sum_j{d_j^{\mathrm{tot}}}$ [km] & 828,931 & 992,453 & 1,062,781 & 1,007,080 \\
Total energy consumption $\sum_jE_j$ [kWh] & 15,215 & 15,147 & 14,610 & 13,227 \\
Monthly operating cost ${C_{\mathrm{total}}}$ [USD] & 56,173 & 76,023 & 98,833 & 105,021 \\
\leavevmode{Operating cost per delivery} $c_{\mathrm{OD}}$ [USD/delivery] & 1.178 & 1.595 & 2.073 & 2.203 \\
Cost per flight $c_F$ [USD/flight] & 1.974 & 1.847 & 1.901 & 1.997 \\
Initial vehicle acquisition cost $C_{\mathrm{acq}}$ [USD] & 49,933 & 58,733 & 61,450 & 62,885 \\
\bottomrule
\end{tabularx}
\end{table}

%% file: tables_ver3/cost_table.tex
% KO4 display values rounded to whole USD from the audited saved results.
% Generated from audited MATLAB results by Re_ExportPaperKO2.m
\begin{table}[!t]
\centering
\caption{Monthly operating cost breakdown [USD] by hub count $H$.}
\label{tab:costparts}
\footnotesize
\setlength{\tabcolsep}{4pt}
\renewcommand{\arraystretch}{1.18}
\begin{tabularx}{\linewidth}{@{}Yrrrr@{}}
\toprule
Cost component & \multicolumn{1}{l}{$H=1$} & \multicolumn{1}{l}{$H=2$} & \multicolumn{1}{l}{$H=3$} & \multicolumn{1}{l@{}}{$H=4$} \\
\midrule
Battery charging & 1,840 & 1,832 & 1,767 & 1,599 \\
Battery replacement & 17,573 & 17,654 & 17,371 & 15,875 \\
Motor replacement & 4,769 & 5,939 & 6,479 & 6,299 \\
Hub operation & 31,991 & 50,598 & 73,215 & 81,247 \\
\bottomrule
\end{tabularx}
\par\smallskip
\parbox{\linewidth}{\scriptsize \leavevmode{Note:} Component sums may differ from the totals in Table~\ref{tab:operation} due to rounding.}
\end{table}

%% file: tables_ver3/comparison_main5_table.tex
% KO4 one-column layout; values preserved from ko2_assets/comparison_main5_table.tex.
% Source: Re_Main5_Results/20261006_143445_871/Re_HubDesignMatrix.mat.
\begin{table}[!t]
\centering
\caption{Monthly operating results by vehicle design $s$ and operating hub count $H$.}
\label{tab:comparison}
\footnotesize
\setlength{\tabcolsep}{3pt}
\renewcommand{\arraystretch}{1.18}
\begin{tabular*}{\linewidth}{@{\extracolsep{\fill}}ccrrrrr@{}}
\toprule
\multicolumn{1}{@{\extracolsep{\fill}}l}{$H$} & \multicolumn{1}{l}{$s$} & \multicolumn{1}{l}{\makecell[l]{Operating cost\\${C_{\mathrm{total}}}$\\{[USD]}}} & \multicolumn{1}{l}{\makecell[l]{{Operating cost}\\{per delivery}\\$c_{\mathrm{OD}}$\\{[USD/delivery]}}} & \multicolumn{1}{l}{\makecell[l]{Flights\\$F$\\{[flights]}}} & \multicolumn{1}{l}{\makecell[l]{Energy\\$\sum_jE_j$\\{[kWh]}}} & \multicolumn{1}{l@{}}{\makecell[l]{Fleet size\\$\sum_jN_j$\\{[drones]}}} \\
\midrule
1 & 1 & 56,173 & \textbf{1.178} & 28,456 & 15,215 & 20 \\
\midrule
2 & 1 & 77,029 & 1.616 & 37,434 & 19,489 & 26 \\
2 & 2 & 76,023 & \textbf{1.595} & 41,169 & 15,147 & 25 \\
\midrule
3 & 1 & 92,322 & \textbf{1.937} & 47,381 & 20,679 & 28 \\
3 & 2 & 96,968 & 2.034 & 49,815 & 16,622 & 28 \\
3 & 3 & 98,833 & 2.073 & 52,003 & 14,610 & 27 \\
\midrule
4 & 1 & 110,974 & 2.328 & 49,323 & 21,251 & 29 \\
4 & 2 & 109,599 & 2.299 & 49,670 & 16,519 & 28 \\
4 & 3 & 105,862 & 2.221 & 51,831 & 14,081 & 27 \\
4 & 4 & 105,021 & \textbf{2.203} & 52,587 & 13,227 & 28 \\
\bottomrule
\end{tabular*}
\par\smallskip
\begin{minipage}{\linewidth}\scriptsize
Note: Bold values indicate the lowest {operating cost per delivery} among the designs evaluated for each $H$.
\end{minipage}
\end{table}